\documentclass{article}

    \usepackage[nonatbib,preprint]{neurips_2023}

\usepackage{tabularx}
\usepackage{multirow}
\usepackage{collcell}

\usepackage{graphics}
\usepackage{caption}

\usepackage{MnSymbol}
\usepackage{algpseudocode}

\usepackage{floatrow}
\newfloatcommand{capbtabbox}{table}[][\FBwidth]
\usepackage{blindtext}
\usepackage{booktabs} 
\usepackage{bbm}

\usepackage{amsmath,amsfonts,bm}

\def\eqref#1{equation~\ref{#1}}

\def\1{\bm{1}}

\DeclareMathAlphabet{\mathsfit}{\encodingdefault}{\sfdefault}{m}{sl}
\SetMathAlphabet{\mathsfit}{bold}{\encodingdefault}{\sfdefault}{bx}{n}

\usepackage{thmtools} 
\usepackage{thm-restate}
\usepackage[utf8]{inputenc} 
\usepackage[T1]{fontenc}    
\usepackage{hyperref}       
\usepackage{url}            
\usepackage{booktabs}       
\usepackage{nicefrac}       
\usepackage{microtype}      
\usepackage{xcolor}         
\usepackage{graphicx}
\usepackage{subfigure}
\usepackage{amsmath}
\usepackage{amsthm}
\usepackage{mathrsfs}
\usepackage{algorithm}
\usepackage{algpseudocode}
\usepackage{verbatim}
\usepackage{cite}
\usepackage{wrapfig}

\newtheorem{remark}{\bf Remark}

\newtheorem{corollary}{\bf Corollary}

\makeatletter
\def\mathcolor#1#{\@mathcolor{#1}}
\def\@mathcolor#1#2#3{%
  \protect\leavevmode
  \begingroup
    \color#1{#2}#3%
  \endgroup
}
\makeatother

\title{End-to-End Supervised Multilabel Contrastive Learning}

\author{
 Ahmad Sajedi\textsuperscript{1}\thanks{Equal contribution~~~$\dag$Corresponding Author},~
 Samir Khaki\textsuperscript{1$*$},~
 Konstantinos N. Plataniotis\textsuperscript{1},~
 Mahdi S. Hosseini\textsuperscript{2$\dag$}\\ 
 \textsuperscript{1}Department of Electrical \& Computer Engineering, University of Toronto, Canada\\
  \textsuperscript{2}Department of Computer Science and Software Engineering, Concordia Univeristy, Canada \\
 {\tt\small \{ahmad.sajedi,samir.khaki\}@mail.utoronto.ca,}\\ {\tt\small kostas@ece.utoronto.ca, mahdi.hosseini@concordia.ca}
}

\begin{document}

\maketitle

\begin{abstract}
Multilabel representation learning is recognized as a challenging problem that can be associated with either label dependencies between object categories or data-related issues such as the inherent imbalance of positive/negative samples. Recent advances address these challenges from model- and data-centric viewpoints. In model-centric, the label correlation is obtained by an external model designs (e.g., graph CNN) to incorporate an inductive bias for training. However, they fail to design an end-to-end training framework, leading to high computational complexity. On the contrary, in data-centric, the realistic nature of the dataset is considered for improving the classification while ignoring the label dependencies. In this paper, we propose a new end-to-end training framework--dubbed KMCL (Kernel-based Mutlilabel Contrastive Learning)--to address the shortcomings of both model- and data-centric designs. The KMCL first transforms the embedded features into a mixture of exponential kernels in Gaussian RKHS. It is then followed by encoding an objective loss that is comprised of (a) reconstruction loss to reconstruct kernel representation, (b) asymmetric classification loss to address the inherent imbalance problem, and (c) contrastive loss to capture label correlation. The KMCL models the uncertainty of the feature encoder while maintaining a low computational footprint. Extensive experiments are conducted on image classification tasks to showcase the consistent improvements of KMCL over the SOTA methods. PyTorch implementation is provided in \url{https://github.com/mahdihosseini/KMCL}.
\end{abstract}

\section{Introduction}
Learning from multilabel representation is a common practice that is considered in both computer vision \cite{everingham2010pascal, chua2009nus, lin2014microsoft, kuznetsova2020open} and medical image \cite{wang2017chestx, hosseini2019atlas, amgad2019structured} application domains. Images usually contain more than one object for classification, where they can be semantically related to each other. The idea is to create an embedded feature space that can capture label dependencies to improve the classification task \cite{chen2019multi, zhao2021transformer, zhu2021residual, ye2020attention}. However, effectively learning such embedded space is known to be a challenging problem and various methods have been proposed over the past few years, including sequence-to-sequence modeling \cite{wang2016cnn, yang2016exploit, chen2018recurrent}, graph approaches \cite{chen2019multi, chen2019learning, ye2020attention, zhao2021transformer, singh2022iml}, and new loss-function designs \cite{ridnik2021asymmetric, ben2022multi, zhang2022use}. Generally, there are two main approaches to addressing the multilabel representation learning problem: the \textit{data-centric} approach and the \textit{model-centric} approach. The data-centric approach focuses on addressing data-related issues like inherent imbalance \cite{ridnik2021asymmetric}, impartial label training \cite{ben2022multi}, and hierarchical relationships \cite{zhang2022use} while ignoring label dependencies. On the contrary, the model-centric approach aims to capture label interactions for semantic embedding such as graph convolutional networks \cite{chen2019multi}, attention mechanisms \cite{ye2020attention}, and transformer-based learning \cite{zhao2021transformer}. Despite the benefits, they fail to design an end-to-end learning framework due to their high computational costs or the laborious task of capturing heuristic label dependencies like using correlation matrices. These limitations make them challenging to implement, optimize, and interpret.

In this paper, we aim to combine the benefits of both data-centric and model-centric approaches while addressing their potential drawbacks. The solution lays on the foundation of asymmetric loss \cite{ridnik2021asymmetric} which tackles the imbalance between positive and negative samples in multilabel classification. Our design augments this loss function by capturing the semantic relationships between labels using a kernel-based contrastive loss. This is achieved through two steps: (a) leveraging a Kernel Mixture Module (KMM) to explore the epistemic uncertainty of the feature encoder (see Figs. \ref{fig:kmcl} and \ref{fig:kmm}). This is done by converting the embedded features of multilabel images into a Gaussian Reproducing Kernel Hilbert Space (RKHS) $\mathcal{H}$, and (b) employing a contrastive learning framework on the Gaussian RKHS to capture label dependencies through a weighted loss-function design (see Fig. \ref{fig:kmcl}). The resulting loss is trainable from end-to-end, providing high numerical stability during training. The following summarizes the contribution of the paper:

\textbf{[C1]}: We propose a novel end-to-end framework --dubbed KMCL-- to strike a balance between model-centric and data-centric approaches using a new contrastive loss augmented on asymmetric classification loss from \cite{ridnik2021asymmetric}. KMCL is capable of capturing both the epistemic uncertainty of the model and label dependencies between classes simultaneously.

\textbf{[C2]}: We introduce a KMM block design within the KMCL framework to generate a mixture of exponential kernels in Gaussian RKHS to model the uncertainty of the feature encoder and improve the robustness of the classification task. To reconstruct the mixture kernels from data, we propose a loss function $\mathcal{L}_{\text{REC}}$ (in Eq. \ref{eq:recon}) as an alternative to the negative log-likelihood loss that addresses the numerical instabilities mentioned in \cite{chen2019face, makansi2019overcoming, zhang2020improved}.

\textbf{[C3]}: We construct the $\mathcal{L}_{\text{KMCL}}$ (in Eq. \ref{eq:kmcl}) as a complementary loss to $\mathcal{L}_{\text{ASL}}$ \cite{ridnik2021asymmetric} to capture label dependencies and enhance classification performance. We utilize the Bhattacharyya coefficient ($\rho$) as a similarity metric between two kernel representations to pull together similar classes (positive) from a pair of multilabel images while contrasting dissimilar ones (negative) in Gaussian RKHS.

\textbf{[C4]}: We consistently improve classification performance on both computer vision and medical imaging tasks with low computational footprints. Our loss design yields robust behavior toward a range of hyperparameters that are fixed across all experiments.

\vspace{-2pt}
\subsection{Related Work}
\textbf{Multilabel Image Representation.} Multilabel image representation problems have been extensively studied, focusing on exploiting label dependencies within semantically aware regions. Previous approaches include RNN-CNN models for sequence-to-sequence modeling \cite{wang2016cnn}, transforming the problem into a multi-instance problem \cite{yang2016exploit}, and using recurrent attention reinforcement learning \cite{chen2018recurrent}. Later, efforts were made to incorporate linguistic embedding of training labels into graph neural network designs \cite{chen2019multi, chen2019learning}. However, graph-based approaches assume the presence of coexisting label dependencies, which may not hold true when labels co-occur infrequently. Attention mechanisms have been introduced in dynamic graph modeling networks to address this issue \cite{ye2020attention, zhao2021transformer}. Despite their effectiveness, these approaches often result in complex models with heavy computational requirements and limited generalization in different domains. A residual attention mechanism was introduced \cite{zhu2021residual} to reduce such complexities by augmenting independent class feature scores using a class-agnostic average pooling method for aggregation scoring. Recent developments in this field emphasize the realistic nature of multilabel data representation. For example, the design proposed in \cite{ridnik2021asymmetric} introduces an asymmetric loss function to balance the frequency of positive and negative classes. Other approaches include class-aware loss design for impartial label training \cite{ben2022multi} and exploring hierarchical relationships of multilabel data in a contrastive learning framework \cite{zhang2022use}. In this paper, we leverage both data- and model-centric approaches to reduce the above-mentioned complexities. 

\textbf{Contrastive Learning.} Self-supervised learning methods primarily focus on contrastive learning, which involves capturing inter-relational object information in image representation. This is achieved through the use of contrastive loss functions, either in unsupervised contrastive learning where labels are absent \cite{chen2020simple, chen2020big, oord2018representation, he2020momentum, cheng2019modified, jaiswal2021survey}, or in supervised contrastive learning where labels are available \cite{khosla2020supervised, chen2021intriguing, malkinski2020multi, finzi2020probabilistic}. The framework has been extended to multilabel representation learning \cite{malkinski2020multi, dao2021multi} by considering shared label images as positive and unshared label images as negative. The existing multilabel contrastive loss designs rely on hard-coded features and lack flexibility in representing semantically aware objects and their label dependencies. However, we propose transforming embedded features into a mixture of exponential kernels in Gaussian RKHS to account for the potential uncertainty of model parameters and accordingly relax the embeddings.
\section{Background on Bhattacharyya Coefficient between Exponential Kernels}\label{sec_Bhattacharyya}
The Bhattacharyya coefficient is a widely used metric to measure the similarity between probability distributions in various fields, including computer vision, pattern recognition, and statistical analysis \cite{patra2015new, pandy2022transferability, hu2021simple, sinha2020neural, combalia2020uncertainty}. Normal distributions are commonly evaluated using this metric to determine class separability in transfer learning \cite{pandy2022transferability}, perform point cloud instance segmentation \cite{zhang2021point}, and employ pseudo-labels for semi-supervised classification \cite{hu2021simple}. However, the Gaussian probability may not always be the best option for estimating the target variable due to normality assumptions which leads to numerical instabilities such as singularity \cite{chen2019face, makansi2019overcoming, zhang2020improved}. A mixture of exponential kernels can be used as a reliable alternative to estimate the relative likelihood of the target variable, especially when the distribution is unknown or multimodal. In such cases, the Bhattacharyya coefficient $\rho$ between the normalized versions of the kernel components can assess the geometric similarity and degree of overlap. Compared to Kullback-Leibler divergence \cite{kullback1951information} or $L_p$ norms, $\rho$ takes values in the range of [0, 1], which makes it a practical choice for comparing two statistical samples. In the following remark, we will elaborate on the closed-form expression of $\rho$ between two exponential kernels.
\begin{remark} \label{theo:NBC}
Let $p(\mathbf{x}):= K_{\boldsymbol{\Sigma}_{p}}(\mathbf{x}, \boldsymbol{\mu}_{p}) = \exp{\left(-\frac{1}{2}\|\mathbf{x} - \boldsymbol{\mu}_{p}\|^{2}_{\boldsymbol{\Sigma}_{p}^{-1}}\right)}$ and $q(\mathbf{x}) := K_{\boldsymbol{\Sigma}_{q}}(\mathbf{x}, \boldsymbol{\mu}_{q}) = \exp{\left(-\frac{1}{2}\|\mathbf{x} - \boldsymbol{\mu}_{q}\|^{2}_{\boldsymbol{\Sigma}_{q}^{-1}}\right)}$ be anisotropic multivariate squared exponential kernels that define a Gaussian RKHS $\mathcal{H}$ \cite{steinwart2006explicit, gretton2013introduction, xu2006explicit}. Then, the Bhattacharyya coefficient between the normalized $p(\mathbf{x})$ and $q(\mathbf{x})$
is:
\vspace*{-0.2cm}
\begin{flalign} \label{eq:BDGen}
\rho\big(p(\mathbf{x}), q(\mathbf{x}) \big)  = \int \Big(\dfrac{p(\mathbf{x})}{\int p(\mathbf{x}) d\mathbf{x}}\Big)^{\frac{1}{2}}\Big(\dfrac{q(\mathbf{x})}{\int q(\mathbf{x}) d\mathbf{x}}\Big)^{\frac{1}{2}}d\mathbf{x} = \frac{\big|\boldsymbol{\Sigma}_{p}\big|^{\frac{1}{4}}\big|\boldsymbol{\Sigma}_{q}\big|^{\frac{1}{4}}}{\big|\boldsymbol{\Sigma}\big|^{\frac{1}{2}}}\exp{\Big(-\frac{1}{8}\|\boldsymbol{\mu}_{p}-\boldsymbol{\mu}_{q}\|^{2}_{\boldsymbol{\Sigma}^{-1}}\Big)},
\end{flalign}

   \vspace{-10pt}
where, $\|\boldsymbol{\mu}_{p}-\boldsymbol{\mu}_{q}\|^{2}_{\boldsymbol{\Sigma}^{-1}} = (\boldsymbol{\mu}_{p}-\boldsymbol{\mu}_{q})^{T}\boldsymbol{\Sigma}^{-1}(\boldsymbol{\mu}_{p}-\boldsymbol{\mu}_{q})$ and $\boldsymbol{\Sigma} = \frac{\boldsymbol{\Sigma}_{p}+\boldsymbol{\Sigma}_{q}}{2}$. The $\boldsymbol{\mu}_{p}, \boldsymbol{\mu}_{q} \in \mathbb{R}^{M}$ and $\boldsymbol{\Sigma}_{p}, \boldsymbol{\Sigma}_{q} \in \mathbb{S}_{++}^{M}$ are the mean vectors and the covariance matrices, respectively, and the operation $|\cdot|$ represents the determinant of a matrix. The proof of Remark \ref{theo:NBC} is provided in Supplementary material. 
\end{remark}
\vspace{-5pt}
The Bhattacharyya coefficient, also known as the Hellinger affinity \cite{kailath1967divergence}, measures the normalized correlation between the square roots of kernels over the entire space. This similarity metric compares $p(\mathbf{x})$ and $q(\mathbf{x})$ by projecting their square roots onto a unit hypersphere and measuring the cosine of the angle between them in the complete inner product space $\mathcal{H}$. 
\begin{figure}
    \centering
    \includegraphics[width=1.0\textwidth]{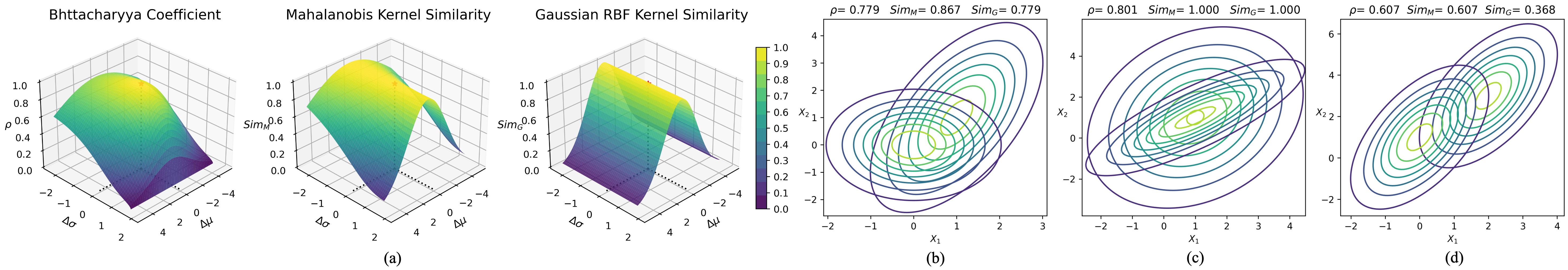}
    \caption{\small (a) 3D plots of similarity measures (Bhattacharyya coefficient $\rho$, Mahalanobis $\text{Sim}_{\text{M}}$, and Gaussian RBF $\text{Sim}_{\text{G}}$) between two 1D kernels based on mean and variance differences (i.e., $\Delta{\mu}$ and $\Delta{\sigma}$). The symbol $\mathcolor{red}{\filledstar}$ denotes the highest value 1 when $\Delta \mu = \Delta \sigma = 0$. Contour plots of 2D kernels are shown with different means and covariance matrices in (b), the same means but different covariance matrices in (c), and different means but the same covariance matrices in (d). Similarity measures $\rho$, $\text{Sim}_{\text{M}}$, and $\text{Sim}_{\text{G}}$ provided for each scenario.\vspace{-15pt}}
    \label{fig:sim}
    \vspace{-9pt}
\end{figure}

   \vspace{-0.15cm}
A careful examination of Equation \ref{eq:BDGen} reveals that the Bhattacharyya coefficient between normalized $p(\mathbf{x})$ and $q(\mathbf{x})$ consists of two terms: a scale factor and an exponential component. The scale factor measures overlap by comparing the generalized variances of the kernels, which are determined by the determinant of their covariance matrices. The scale factor converges to one when the covariance matrices of the two kernels are similar, indicating an overlap between them. The generalized variance of a kernel is related to its entropy and power entropy \cite{Cover2006}, which measure uncertainty and spread. This allows the scale factor to consider differences in information content and orientation, resulting in separability due to covariance dissimilarity. On the other hand, the second term measures the similarity between the means $\boldsymbol{\mu}_{p}$ and $\boldsymbol{\mu}_{q}$ weighted by the precision matrix $\boldsymbol{\Sigma}^{-1}$, providing separability based on positional differences. This exponential component represents the Mahalanobis kernel similarity \cite{herbrich2001learning} between $\boldsymbol{\mu}_{p}$ and $\boldsymbol{\mu}_{q}$ with respect to $\boldsymbol{\Sigma}^{-1}$. The following corollary will further elucidate the connection of the Bhattacharyya coefficient with the Mahalanobis and Gaussian similarities. 
\begin{corollary}\label{corr1}
Let $p(\mathbf{x}) := K_{\boldsymbol{\Sigma}_{p}}(\mathbf{x}, \boldsymbol{\mu}_{p})$ and $q(\mathbf{x}) := K_{\boldsymbol{\Sigma}_{q}}(\mathbf{x}, \boldsymbol{\mu}_{q})$ be multivariate kernels defined in Remark \ref{theo:NBC}. The Bhattacharyya coefficient between normalized $p(\mathbf{x})$ and $q(\mathbf{x})$ can be reduced to either the Mahalanobis or the RBF kernel similarity, depending on the covariance matrices:\vspace{+3pt}\\
     (i) The \textbf{Mahalanobis kernel similarity}, $\text{Sim}_{\text{M}}{\big(p(\mathbf{x}), q(\mathbf{x})\big)}$, is obtained when the covariance matrices are homoscedastic, i.e., $\boldsymbol{\Sigma}_{p} = \boldsymbol{\Sigma}_{q} = \boldsymbol{\Sigma}$. It has the following closed-form expression: 
     \vspace*{-0.17cm}
\begin{subequations}
  \begin{flalign} \label{eq:MK}
  \text{Sim}_{\text{M}}{\big(p(\mathbf{x}), q(\mathbf{x})\big)} = \rho\big(K_{\boldsymbol{\Sigma}}(\mathbf{x}, \boldsymbol{\mu}_{p}), K_{\boldsymbol{\Sigma}}(\mathbf{x}, \boldsymbol{\mu}_{q})\big) =  \exp\big(-\dfrac{1}{2 (2)^{2}}\big\|\boldsymbol{\mu}_{p}-\boldsymbol{\mu}_{q}\big\|^{2}_{\boldsymbol{\Sigma}^{-1}}\big).
  \end{flalign}

    \vspace{-9pt}
  The described Mahalanobis metric evaluates the similarity between $p(\mathbf{x})$ and $q(\mathbf{x})$ based on their mean difference and relative positions (see Fig. \ref{fig:sim}d).\vspace{+3pt}\\
  (ii) The \textbf{Gaussian kernel similarity}, ${\text{Sim}_{\text{G}}(p(\mathbf{x}), q(\mathbf{x}))}$, is obtained when the covariance matrices are equal and isotropic, meaning $\boldsymbol{\Sigma}_{p} = \boldsymbol{\Sigma}_{q} = \sigma^{2}\boldsymbol{I}$. The closed-form expression will be:\vspace*{-0.15cm}
   \begin{flalign} \label{eq:GK}
   \text{Sim}_{\text{G}}(p(\mathbf{x}), q(\mathbf{x})) = \rho\big(K_{\boldsymbol{\Sigma}}(\mathbf{x}, \boldsymbol{\mu}_{p}), K_{\boldsymbol{\Sigma}}(\mathbf{x}, \boldsymbol{\mu}_{q})\big) = \exp\big(-\dfrac{\|\boldsymbol{\mu}_{p}-\boldsymbol{\mu}_{q}\|^{2}}{8\sigma^{2}}).
   \end{flalign}
\end{subequations}
\end{corollary}

   \vspace{-7pt}
In cases where two kernels have similar means but different covariance matrices, the Mahalanobis and Gaussian kernel similarities often exhibit a perfect correlation that may not precisely reflect true similarities (Figs. \ref{fig:sim}a and c). Instead, the Bhattacharyya coefficient evaluates the generalized variances of the kernels and identifies similarities in their orientation, shape, and means (Figs. \ref{fig:sim}a and c). Therefore, it is often a superior metric to the Mahalanobis and the Gaussian kernel similarities.

\vspace{-3pt}
The process of computing the final value of the closed-form expression between high-dimensional kernels can be time-consuming and resource-intensive. This problem can be alleviated by imposing constraints on the mean vectors and/or the covariance matrices. Following \cite{ding2015articulated, park2022probabilistic}, we will cover how specific constraints can be applied to improve computational efficiency in a subsequent corollary.

\begin{corollary} \label{cor2}
Let $p(\mathbf{x}) := K_{\boldsymbol{\Sigma}_{p}}(\mathbf{x}, \boldsymbol{\mu}_{p})$ and $q(\mathbf{x}) := K_{\boldsymbol{\Sigma}_{q}}(\mathbf{x}, \boldsymbol{\mu}_{q})$ be two multivariate kernels as defined in Remark \ref{theo:NBC}. The following statements hold:\vspace{+2pt}\\
(i) If the covariance matrices are diagonal, meaning that 
$\boldsymbol{\Sigma}_{p} = \text{diag}(\sigma_{p,1}^{2}, \cdots, \sigma_{p,M}^{2})$ and $\boldsymbol{\Sigma}_{q} = \text{diag}(\sigma_{q,1}^{2}, \cdots, \sigma_{q,M}^{2})$, the Bhattacharyya coefficient between normalized $p(\mathbf{x})$ and $q(\mathbf{x})$ will be
\begin{subequations}
\begin{flalign}
\rho(p(\mathbf{x}), q(\mathbf{x})) = \Big(\prod_{i=1}^{M}\big(\frac{\sigma_{p,i}^{2}+\sigma_{q,i}^{2}}{2\sigma_{p,i}\sigma_{q,i}}\big)^{-\frac{1}{2}}\Big)\exp\Big(-\dfrac{1}{4}\sum_{i=1}^{M}\frac{({\mu}_{p,i} -{\mu}_{q,i})^{2}}{\sigma_{p,i}^{2}+\sigma_{q,i}^{2}}\Big). \hspace{0.75cm} (\textit{Anisotropic})
\end{flalign}
(ii) If the mean vectors have identical values across all dimensions $(\boldsymbol{\mu}_{p} = \mu_{p}\mathbf{1}, \boldsymbol{\mu}_{q} = \mu_{q}\mathbf{1}$, where $\mathbf{1} = [1, \cdots, 1]^{T}\in \mathbb{R}^{M}$ is the one vector$)$, and the covariance matrices are diagonal with homogeneous variances $(\boldsymbol{\Sigma}_{p} = \sigma_{p}^{2}\boldsymbol{I}$, $\boldsymbol{\Sigma}_{q} = \sigma_{q}^{2}\boldsymbol{I}$, where $\boldsymbol{I}\in{\mathbb{S}^{M}_{++}}$ is the identity matrix$)$, then the Bhattacharyya coefficient between two normalized isotropic kernels $p(\mathbf{x})$ and $q(\mathbf{x})$ can be calculated as
\begin{flalign} \label{eq:BD}
\rho(p(\mathbf{x}), q(\mathbf{x})) =  \Big(\frac{\sigma_{p}^{2}+\sigma_{q}^{2}}{2\sigma_{p}\sigma_{q}}\Big)^{-\frac{M}{2}}\exp\Big(-\dfrac{M}{4}\frac{({\mu}_{p} -{\mu}_{q})^{2}}{\sigma_{p}^{2}+\sigma_{q}^{2}}\Big).   \hspace{0.75cm} (\textit{Isotropic})
\end{flalign}
\end{subequations}
\end{corollary}

\section{Proposed Method} \label{sec:kmcl}
\begin{wrapfigure}[13]{R}{0.68\linewidth}
    \centering
    \vspace{-2.4 \intextsep}
    \hspace*{-0.3\columnsep}\includegraphics[width=1
    \linewidth]{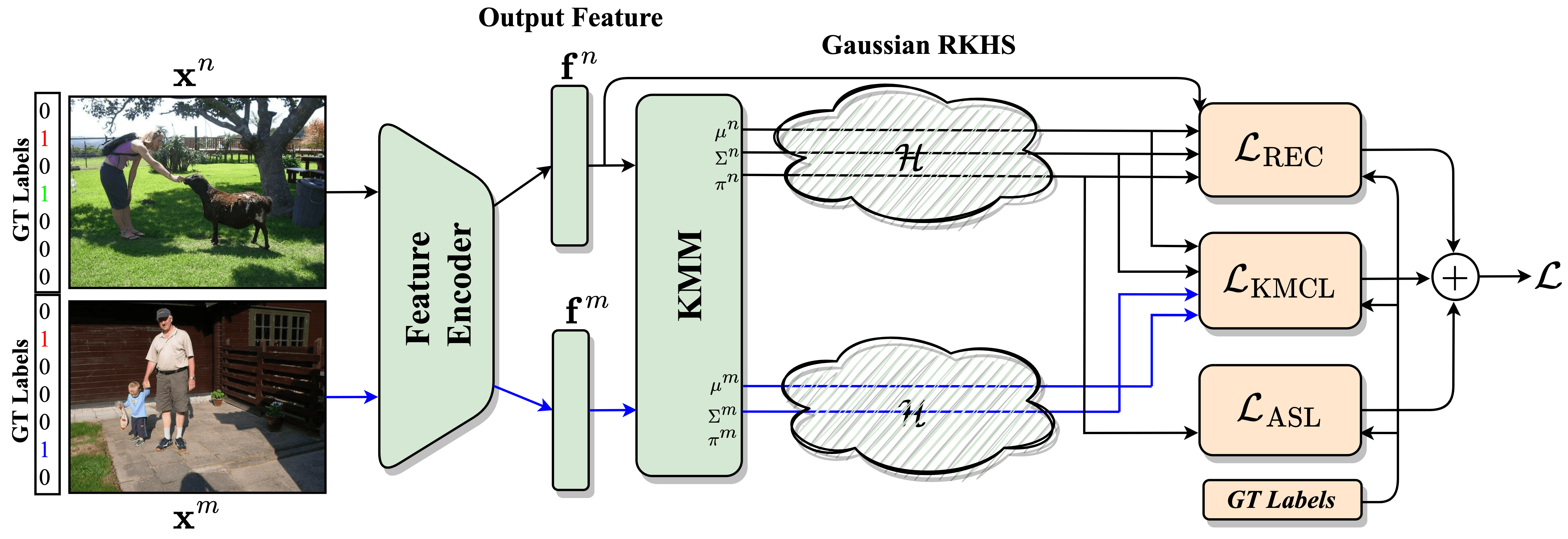}
    \caption{\textbf{Overview of KMCL framework.} The training pipeline comprises a feature encoder that feeds into the KMM, which outputs the parameters of a mixture model in the Gaussian RKHS $\mathcal{H}$. These parameters then define the objective function that captures label correlation to aid in training the model for the multi-label classification.\vspace{-12pt}}
    \label{fig:kmcl}
    \vspace{-6pt}
\end{wrapfigure}

The multi-label classification task involves assigning multiple labels to an image $\mathbf{x}^{n}$ from sample space $\mathbf{X}$. These labels are typically correlated with each other and represented by a multi-hot binary vector $\mathbf{y}^{n} \in \{0,1\}^{K}$, where $K$ denotes the number of labels. In this section, we propose an end-to-end multi-label learning framework--dubbed Kernel-based multi-label Contrastive Learning (KMCL), that captures label correlations to improve recognition performance. Given an input batch of data, we first propagate it through the encoder network to obtain the feature embedding. The embedding is then inputted into a novel fully connected layer called the Kernel Mixture Module (KMM), which produces a Gaussian Reproducing Kernel Hilbert Space $\mathcal{H}$. The Gaussian RKHS embedding can handle higher-order statistics of the features and has a complete inner product that enables linear geometry, making it richer than the deterministic feature embedding. Finally, we compute the loss function using the KMM outputs on space $\mathcal{H}$ to capture label correlation and train the model for multi-label classification. Figure \ref{fig:kmcl} provides a visual explanation.

\subsection{KMCL Framework} \label{subsec:framework}
The main components of the KMCL framework are:
\begin{wrapfigure}[2]{R}{0.23\linewidth}
    \centering
    \vspace{-4.2\intextsep}
    \hspace*{-1\columnsep}\includegraphics[width=1.06
    \linewidth]{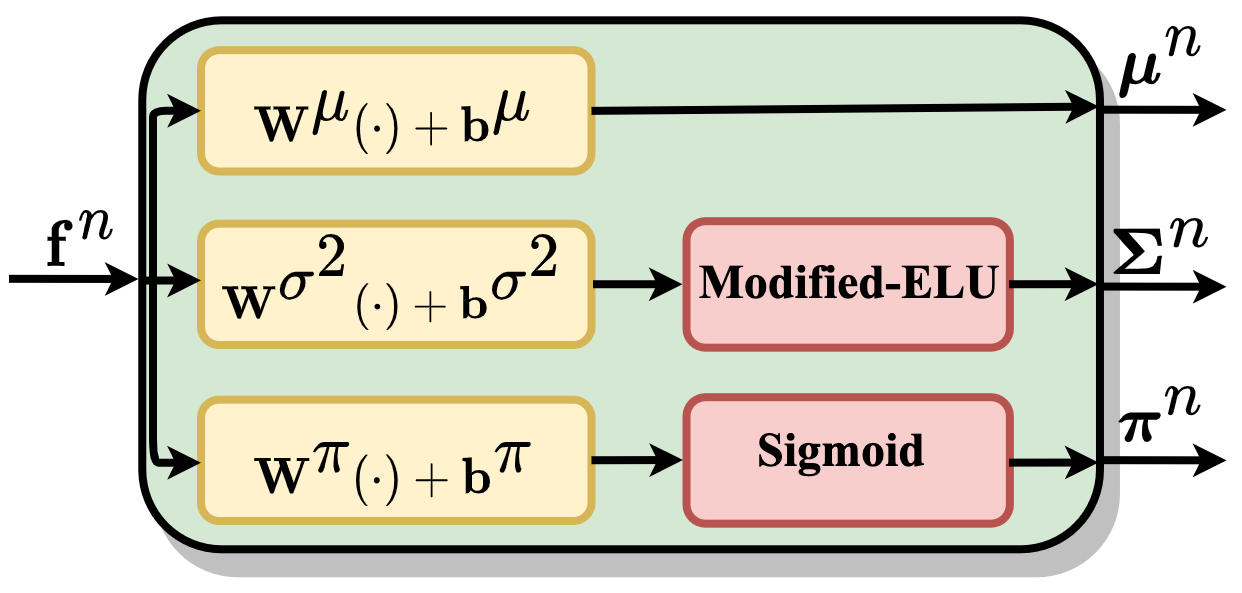}
    \caption{\textbf{Internal architecture of KMM.}}
    \label{fig:kmm}
    \vspace{-8pt}
\end{wrapfigure}
\vspace{-6pt}
\paragraph{Feature Encoder.}The encoder network takes two samples from the input batch separately and generates corresponding feature representation vectors $\mathbf{f} \in \mathbb{R}^M$. The dimension of the feature vector depends on the encoder type. 
\vspace{-6pt}
\paragraph{KMM.} 
Most feature encoders produce deterministic results that do not quantify or control uncertainty, leading to low confidence in robust multi-label classification tasks and errors in interpreting the output predictions. Uncertainty in deep learning arises from two sources: epistemic uncertainty (model uncertainty), resulting from uncertainty in model parameters, and aleatoric uncertainty (data uncertainty), which stems from the inherent noise in data and label ambiguity. In this study, we propose the Kernel Mixture Module (KMM) to estimate epistemic uncertainty in predictions. The KMM takes the feature vector $\mathbf{f}$ from the encoder network and generates a mixture of exponential kernels within the Hilbert space, each corresponding to a specific class in an image. Specifically, the fully connected layer in the KMM utilizes learnable weights and biases to produce three outputs for each unimodal exponential kernel component: the mixture coefficient $\pi_{k}$, mean vector $\boldsymbol{\mu}_{k}$, and covariance matrix $\boldsymbol{\Sigma}_{k}$ (Fig. \ref{fig:kmm}). The parameters $\pi_{k}$, $\boldsymbol{\mu}_{k}$, and $\boldsymbol{\Sigma}_{k}$ quantify the existence, relative spatial positioning, and relative statistical complexities (measures of spread and uncertainty) of the $k$th class membership. These parameters are then used to model the label representation of a given sample $\mathbf{x}^{n}$ associated with a class vector $\mathbf{y}^{n}$ using the following expression:
\vspace{-6pt} 
\begin{equation}
   \label{eq:kmm}
    \mathcal{G}_{\mathcal{S}}(\mathbf{f}^{n}) := \sum_{k \in \mathcal{S}}\pi_k^{n} g_{k}(\mathbf{f}^{n})  =  
    \sum_{k \in \mathcal{S}} \pi_k^{n} \exp\Big(-\frac{\lVert \mathbf{f}^{n} - {\mu}_{k}^{n}\boldsymbol{1} \rVert^{2}}{2{(\sigma_{k}^{n})}^{2}}\Big),
 \end{equation}

\vspace{-10pt} 
where, $\mathcal{S} = \{k: y_{k}^{n} = 1\}$ and $\mathbf{f}^{n}$ is the extracted feature vector of the input sample. The component $g_{k}(\mathbf{f}^{n}) := K_{\boldsymbol{\Sigma}_{k}^{n}}(\mathbf{f}^{n}, \boldsymbol{\mu}_{k}^{n})$ is an isotropic exponential kernel where $\boldsymbol{\mu}_{k}^{n} = \mu_{k}^{n} \mathbf{1}$, $\boldsymbol{\Sigma}_{k}^{n} = (\sigma^{n}_{k})^{2}\boldsymbol{I}$, and $\pi_{k}^{n} \in [0, 1]$. These adaptive parameters i.e.,  $\boldsymbol{\theta}_{k}^{n} = [\mu_{k}^{n}, (\sigma^{n}_{k})^{2}, \pi_{k}^{n}]$ are calculated through forward propagation, using suitable activation functions to ensure that the parameters adhere to their constraints. The {sigmoid} activation function is used to normalize the mixture coefficient for efficient multi-label classification, accurately predicting the likelihood of multiple labels. The modified version of the exponential linear unit (ELU) \cite{clevert2015fast} is also used as an activation function for variances, ensuring their semi-positivity. The detailed architecture of KMM can be found in Fig. \ref{fig:kmm} and Supplementary material. 
\subsection{Multi-label Learning with KMCL} \label{subsec:loss}
Building upon the KMCL framework, we aim to provide insights into the learning process of multi-label tasks. To achieve this, we introduce the details of our objective function, which comprises three components: \textit{reconstruction loss}, \textit{classification loss}, and \textit{contrastive loss}. Throughout this paper, we use $N$ and $K$ to denote the mini-batch size and the total number of classes, respectively.
\paragraph{Reconstruction Loss.}
\begin{wrapfigure}[12]{R}{0.3\linewidth}
    \centering
    \vspace{-1.4\intextsep}
    \hspace*{-0.5\columnsep}\includegraphics[width=1.06
    \linewidth]{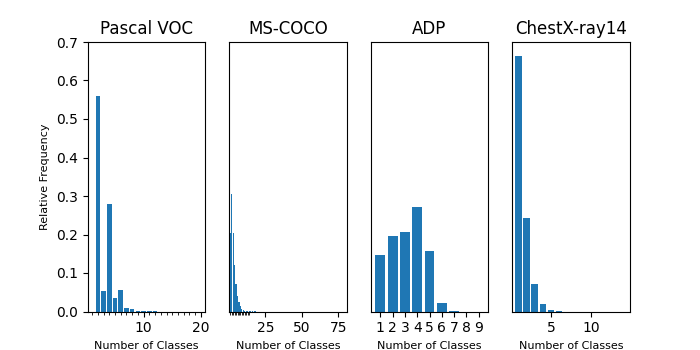}
    \caption{{Relative frequency histograms of class distributions in four datasets show that most images have $2$, $2$, $4$, and $1$   labels in the Pascal-VOC, MS-COCO, ADP, and  ChestX-ray14, respectively.}}
    \label{fig:histo}
    \vspace{-8pt}
\end{wrapfigure}
It is straightforward to compute the mixture model defined in Equation \ref{eq:kmm} using the KMM output parameters, which provide $3K$ values for each input sample. Following this calculation, the model can be used to learn label-level representations in Hilbert space $\mathcal{H}$ by minimizing its negative log-likelihood. Therefore, we introduce to optimize the following reconstruction loss over the data batch to train the mixture model
\begin{flalign} \label{eq:recon}
    \mathcal{L}_{\text{REC}} = \frac{1}{N}\sum_{n=1}^{N}-\log\frac{\mathcal{G}_{\mathcal{S}}(\mathbf{f}^{n})}{\mathcal{G}_{\mathcal{Y}}(\mathbf{f}^{n})},
\end{flalign}
where, $\mathcal{G}_{\mathcal{Y}}(\mathbf{f}^{n}) := \sum_{k\in\mathcal{Y}=\{1, \cdots, K\}}\pi_{k}g_{k}$ and ${\mathcal{G}_{\mathcal{S}}\big(\mathbf{f}^{n})}$ denotes the kernel mixture associated with image $\mathbf{x}^n$ defined in Equation \ref{eq:kmm}. The log-ratio term in Equation \ref{eq:recon} is always negative i.e. $\mathcal{G}_{\mathcal{S}}(\mathbf{f}^{n})\leq \mathcal{G}_{\mathcal{Y}}(\mathbf{f}^{n})$, where the loss is led by the supervised labels for reconstruction. We propose this as an alternative choice for reconstruction loss, which is commonly used in the literature \cite{li2019generating, varamesh2020mixture, peharz2020einsum}. Our new loss function $\mathcal{L}_{\text{REC}}$ exhibits robust behavior without relying on numerical tricks for stabilization.
\paragraph{Classification Loss.}
The analysis in Figure \ref{fig:histo} reveals that despite varying statistical and conceptual properties across datasets, most images have only a fraction of labels, causing a significant imbalance between positive and negative samples. This imbalance can lead to poor training accuracy as gradients from positive labels may be underemphasized. To mitigate this issue, we use ASL \cite{ridnik2021asymmetric} as a classification loss function that adjusts the contributions of positive and negative samples by down-weighting easy negative samples and focusing on the hard ones. Therefore, given the predictive mixture of coefficients $\boldsymbol{\pi}^{n}$ from KMM and the ground-truth multi-hot label vector $\mathbf{y}^{n}$, the classification loss for a batch is obtained as
 \begin{equation}
     \label{eq:asl}
     \mathcal{L}_{\text{ASL}} = \frac{1}{N}\sum_{n=1}^{N} \sum_{k=1}^K -y_{k}^{n}(L_{k}^{n})_{+}-(1-y_{k}^{n})(L_{k}^{n})_{-},
 \end{equation}
where, $(L_{k}^{n})_{+} = (1-\pi_{k}^{n})^{\gamma_{+}}\log (\pi_{k}^{n})$, and $(L_{k}^{n})_{-} = (\max(\pi_{k}^{n}-m, 0))^{\gamma_{-}}\log (1-\max(\pi_{k}^{n}-m, 0))$ represent the positive and negative loss parts, respectively, such that $\gamma_{+}$, $\gamma_{-}$, and $m$ are the hyper-parameters used to balance the loss. For additional information on $\mathcal{L}_{\text{ASL}}$, please refer to \cite{ridnik2021asymmetric}.
\vspace{-4pt}
\paragraph{Kernel-based Contrastive Loss.}
The ASL loss function classifies labels independently, making it difficult to capture correlations between co-occurring semantic labels. Moreover, it fails to account for uncertainty in predictions, which can undermine decision-making confidence. To address these limitations, we propose a new loss function, $\mathcal{L}_{\text{KMCL}}$, which incorporates label correlation and epistemic uncertainty into supervised contrastive learning to improve representation.

The objective of kernel-based multi-label contrastive loss $\mathcal{L}_{\text{KMCL}}$ is to pull together the kernel representations of positive images that have shared classes with the anchor image $\mathbf{x}^{n}$ in the embedding space $\mathcal{H}$, while pushing apart negative samples that do not share any classes. This approach differs from deterministic supervised contrastive losses \cite{khosla2020supervised, zhang2022use, dao2021multi} as $\mathcal{L}_{\text{KMCL}}$ constructs the positive and negative pairs using similarity measures that consider the uncertainty of kernel representations. The similarity is measured by a Bhattacharyya coefficient discussed in Corollary \ref{cor2} (isotropic), which determines the overlap between these exponential kernels and their confidence in proximity. Essentially, the kernel-based contrastive loss optimizes the similarity of frequently co-occurring labels and captures their statistical dependencies, making it a valuable complement to ASL. The contrastive loss is defined for the entire minibatch as follows:
\begin{flalign} \label{eq:kmcl}
\mathcal{L}_{\text{KMCL}} = \frac{1}{N}\sum_{n=1}^{N} \frac{-1}{|\mathcal{A}(n)|} \sum_{m\in  \mathcal{A}(n)}J(n, m) \bigg(\sum_{k\in \mathcal{K}(n, m)}  \log \frac{\exp{\big({\rho_{k,k}^{n,m}/\tau}\big)}}{\sum\limits_{i\in\{N\backslash n\}}\exp{\big({{\rho_{k,k}^{n,i}/\tau}}\big)}}\bigg),
\end{flalign}
where, $\rho_{k,l}^{n,m}:=\rho\big(g_{k}(\mathbf{f}^{n}), g_{l}(\mathbf{f}^{m})\big)$ indicates the Bhattacharyya coefficient between the normalized exponential kernels $g_{k}(\mathbf{f}^{n})$ and $g_{l}(\mathbf{f}^{m})$ (see Corollary \ref{cor2}) and $\tau$ is the temperature parameter. The positive set $\mathcal{A}(n) = \{m \in \{N \backslash n\}: \mathbf{y}^{n} \boldsymbol{\cdot} \mathbf{y}^{m} \neq 0, \text{where $\boldsymbol{\cdot}$ is a dot product.}\}$ includes samples that share at least one label with the anchor image $\mathbf{x}^{n}$, while $\mathcal{K}(n,m)= \{k\in \mathcal{Y}: y_{m}^{k} = y_{n}^{k} = 1\}$ represents the indices of shared labels between $\mathbf{x}^{n}$ and $\mathbf{x}^{m}$. The Jaccard index $J(n, m)=\frac{\mathbf{y}^{n} \boldsymbol{\cdot} \mathbf{y}^{m}}{\|\mathbf{y}^{n}\|^{2}+\|\mathbf{y}^{m}\|^{2}-\mathbf{y}^{n}\boldsymbol{\cdot}\mathbf{y}^{m}}$ serves as a weighting factor for positive samples based on the number of shared labels with the anchor. It measures the intersection over union (IOU) of the label vectors between the anchor and positive image, taking into account object co-occurrences. In this way, $\mathcal{L}_\text{KMCL}$ prioritizes positive samples with a high Jaccard index for a given anchor while downplaying samples with few shared labels.
  \begin{wrapfigure}[14]{R}{0.44\linewidth}
    \centering
    \vspace{-1.2pt}
    \hspace*{-1.0\columnsep}\includegraphics[width=1.06\linewidth]{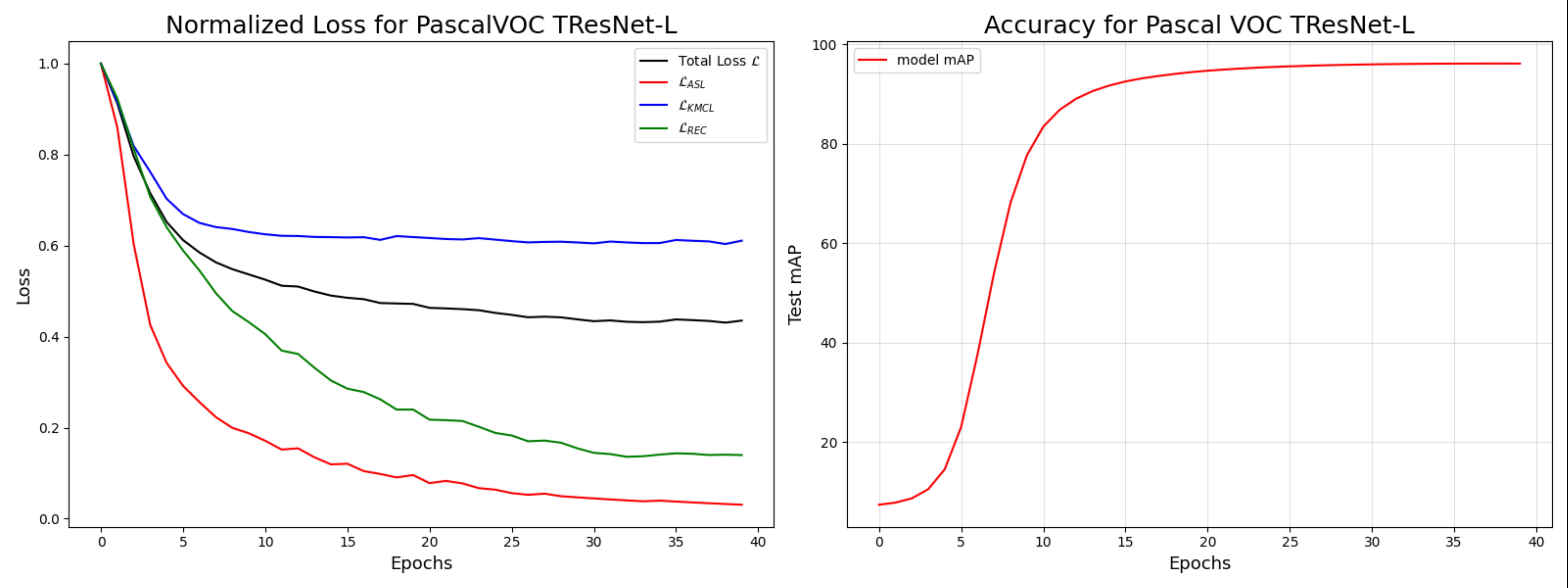}
    \caption{(a) Training loss over different epoch training. Plots show the normalized total loss $\mathcal{L}$ as well as different normalized sub-losses, and (b) Training accuracy of KMCL pipeline over different epoch training}
    \label{fig:losses}
    \vspace{-8pt}
\end{wrapfigure}
\vspace{-4pt}
\paragraph{Objective Function.} The overall training loss of the KMCL is the augmented Lagrangian of the three aforementioned losses, which can be expressed as:
\vspace{-1pt}
\begin{flalign}
    \label{eq:loss}
    \mathcal{L} = \mathcal{L}_{\text{REC}}   + \lambda_1 \mathcal{L}_{\text{ASL}}  + \lambda_2 \mathcal{L}_{\text{KMCL}},
\end{flalign}

\vspace{-4.5pt}
where $\lambda_{1}$ and $\lambda_{2}$ are the Lagrangian multipliers used to balance the gradients of $\mathcal{L}_\text{ASL}$ and $\mathcal{L}_\text{KMCL}$, respectively. We use an end-to-end pipeline to incorporate contrastive learning into supervised classification, which simultaneously trains the feature encoder and classification parts. This approach is different from previous methods that use contrastive losses \cite{chen2020simple, khosla2020supervised, bai2022gaussian}. In those methods, the encoder is trained with a contrastive loss and then frozen before being transferred to the classifier for tuning. Instead, the KMCL framework combines these training regimes into one formulation, enabling us to learn multi-label classification and label correlations with data-driven techniques.
\subsection{KMCL Algorithm}\label{kmcl_alg}
\begin{wrapfigure}[15]{R}{0.57\textwidth}
 \vspace{-2.4\intextsep}
\begin{algorithm}[H]
\caption{Pseudo-code of the KMCL framework}
\label{alg:kmcl}
\begin{algorithmic}[1]
\Require Training set $\{\mathbf{x}^{n}, \mathbf{y}^{n}\}_{n=1}^{N}$, Batch size $N$, epochs
\For{$i = 1, 2, \cdots, epochs$}  \label{line:1}
	\State Sample a mini-batch $\{\mathbf{x}^{n}\}_{n=1}^{N}$ from Training set 
	\For{each $\mathbf{x}^{n}$}
	\State Obtain positive pair $m\in \mathcal{A}(n)$ 
	\State $\mathbf{f}^{n}, \mathbf{f}^{m} =$ Feature Encoder ($\mathbf{x}^{n}, \mathbf{x}^{m}$)
	\State $\boldsymbol{\theta}^{n}, \boldsymbol{\theta}^{m} =$ KMM ($\mathbf{f}^{n}, \mathbf{f}^{m}$) \label{line:6}
	\State Calculate $\mathcal{L}_{\text{REC}}, \mathcal{L}_{\text{ASL}},$ and $\mathcal{L}_{\text{KMCL}}$ (Eq. \ref{eq:recon} - \ref{eq:kmcl}) \label{line:7}
	\State $\mathcal{L} = \mathcal{L}_{\text{REC}}   + \lambda_1 \mathcal{L}_{\text{ASL}}  + \lambda_2 \mathcal{L}_{\text{KMCL}}$ (Eq. \ref{eq:loss}) \label{line:8}
	\State Update the parameters by Backpropagation\label{line:9} 
	\EndFor
\EndFor
\Ensure \text{Final loss value $\mathcal{L}$} 
\end{algorithmic}
\end{algorithm}
  \end{wrapfigure}

The pseudo-code of the proposed KMCL framework is outlined in Algorithm \ref{alg:kmcl}, which takes a set of batches and a specified number of epochs as inputs. The pair of anchor images and their positive set are fed through the network depicted in Figure \ref{fig:kmcl} to obtain the feature vectors and parameters of the corresponding kernel mixtures (lines \ref{line:1}-\ref{line:6}). The overall loss is then computed as an augmented Lagrangian of the $\mathcal{L}_{\text{REC}}$, $\mathcal{L}_{\text{ASL}}$, and $\mathcal{L}_{\text{KMCL}}$ using the KMM parameters (lines \ref{line:7}-\ref{line:8}). Finally, the objective function is back-propagated through the KMM and the feature encoder for each iteration to update the weights based on the gradients associated with the subsequent forward pass (line \ref{line:9}). This iterative process continues until convergence is reached.

Figures \ref{fig:losses} (a) and (b) demonstrate the results of implementing the KMCL framework with TResNet-L \cite{ridnik2021tresnet} as the encoder network on the Pascal-VOC dataset \cite{everingham2010pascal}. Fig. \ref{fig:losses} (b) displays the objective loss behavior along with the evolution of the three loss terms for the training and test sets; whereas The mean average precision (mAP) accuracy is presented in \ref{fig:losses} (a). The losses decrease with different multiplicative factors due to the tuned Lagrangian multipliers. The convergence speed of the method on multi-label tasks is impressive, reaching 96.2$\%$ mAP accuracy in fewer than 30 epochs.
\section{Experiments}
In this section, we present the experimental setup and demonstrate the superior performance of KMCL in both general computer vision and medical imaging domains. To ensure robust feature extraction, we utilized TResNet-M and TResNet-L \cite{ridnik2021tresnet}, state-of-the-art architectures designed for different image resolutions (224 and 448, respectively). The features are then passed through the KMM to obtain the mixture parameters $\boldsymbol{\pi}$, $\boldsymbol{\mu}$, and $\boldsymbol{\Sigma}$. Additional information regarding the encoders, KMM, datasets, evaluation metrics, and training details can be found in Supplementary material. 


\textbf{Datasets.} We evaluate the KMCL's performance on popular computer vision datasets, PASCAL-VOC \cite{everingham2010pascal} and MS-COCO \cite{lin2014microsoft}, as well as on medical datasets, ADP \cite{hosseini2019atlas} and ChestX-ray14 \cite{wang2017chestx}. 

\textbf{Evaluation Metrics.} \label{sec: PM}
Following SOTA \cite{chen2019multi,ridnik2021asymmetric,zhao2021transformer}, we report the standard metrics of mean average precision (mAP), average overall precision (OP), recall (OR), and F1 score (OF1) in addition to per-class precision (CP), recall (CR), and F1 score (CF1). We considered the number of parameters (M) and GMAC as measures of computational costs. Finally, for the ChestX-ray14 dataset \cite{wang2017chestx}, we reported per-class AUC scores to assess model discriminability for specific classes.


\textbf{Training Details.}
We implemented the KMCL framework using PyTorch, following Alg. \ref{alg:kmcl}. The backbone feature encoders were initialized with pre-trained architectures, while the mixture parameters were initialized by applying a uniform distribution to $\boldsymbol{\pi}$ and $\boldsymbol{\mu}$ and setting $\boldsymbol{\Sigma}$ to a constant value of 1. In all experiments, we assign fixed values of 0.1 and 0.3 to $\lambda_1$ and $\lambda_2$ respectively, as specified in Eq. \ref{eq:loss}. The Adam optimizer \cite{kingma2014adam} was used with an initial learning rate of $2e-4$, and the OneCycleLR scheduler \cite{smith2019super} for 40 epochs. Standard augmentations from RandAugment policy \cite{cubuk2020randaugment} were applied to the training data. Experiments were conducted on four NVIDIA GeForce RTX 2080Ti GPUs. 

\textbf{How does KMCL compare to SOTA methods on computer vision datasets?} We evaluate KMCL with SOTA methods on computer vision datasets in Table \ref{table: VOC2007 benchmark} and Fig. \ref{fig:mscoco}. KMCL outperforms the best competitors on PascalVOC and MS-COCO, achieving superior performance with a margin of $0.4\%$ and $0.2\%$ in mAP score, respectively. In particular, KMCL excels in challenging classes on PascalVOC, such as the \textit{sofa} and \textit{bus} classes, with an improvement of over $3.0\%$. On MS-COCO, KMCL demonstrates significant improvements across multiple metrics, including mAP, OF1, and CF1. Using the TResNet-M encoder at resolution 224, we achieve state-of-the-art results with a $5.0\%$ increase in mAP compared to the best method. Similarly, with TResNet-L at a resolution of 448, KMCL surpasses other methods in overall and per-class metrics. These achievements are attained by integrating the proposed contrastive learning with ASL classification loss, to capture label correlation and enhance prediction accuracy. This is illustrated through the Top3-metrics on MS-COCO, where our 3 classes are better selected by considering label correlation when ranking the predictions.

\begin{table}[H]
\vspace{-0.5em}
\tiny
\setlength{\abovecaptionskip}{0pt}
 \caption{\small Comparisons with state-of-the-art methods on the Pascal VOC \cite{everingham2010pascal} dataset. VGG16 and ResNet101/TResNet backbones denoted as (*) and $(\dagger)$ where applicable.} 
	\label{table: VOC2007 benchmark}
 \centering{
\setlength\tabcolsep{3pt}
	\resizebox{\textwidth}{!}{

\begin{tabular}{c|c|c|c|c|c|c|c|c|c|c|c|c|c|c|c|c|c|c|c|c||c}

\specialrule{2pt}{1pt}{1pt}
Methods &
  aero &
  bike &
  bird &
  boat &
  bottle &
  bus &
  car &
  cat &
  chair &
  cow &
  table &
  dog &
  horse &
  motor &
  person &
  plant &
  sheep &
  sofa &
  train &
  tv &
  mAP \\ \hline \hline
CNN-RNN* \cite{wang2016cnn} & 96.7 & 83.1 & 94.2 & 92.8 & 61.2 & 82.1 & 89.1 & 94.2 & 64.2 & 83.6 & 70.0 & 92.4 & 91.7 & 84.2 & 93.7 & 59.8 & 93.2 & 75.3 & \textbf{99.7} & 78.6 & 84.0 \\
Fev+Lv* \cite{yang2016exploit} & 97.9 & 97.0 & 96.6 & 94.6 & 73.6 & 93.9 & 96.5 & 95.5 & 73.7 & 90.3 & 82.8 & 95.4 & 97.7 & 95.9 & 98.6 & 77.6 & 88.7 & 78.0 & 98.3 & 89.0 & 90.6 \\
HCP* \cite{wei2015hcp}  & 98.6 & 97.1 & 98.0 & 95.6 & 75.3 & 94.7 & 95.8 & 97.3 & 73.1 & 90.2 & 80.0 & 97.3 & 96.1 & 94.9 & 96.3 & 78.3 & 94.7 & 76.2 & 97.9 & 91.5 & 90.9 \\
ResNet-101 $\dagger$ \cite{he2016deep} &
  99.8 &
  98.3 &
  98.0 &
  98.0 &
  79.5 &
  93.2 &
  96.8 &
  97.7 &
  79.9 &
  91.0 &
  86.6 &
  98.2 &
  97.8 &
  96.4 &
  98.8 &
  79.4 &
  94.6 &
  82.9 &
  99.1 &
  92.1 &
  92.9 \\
RNN-Attention* \cite{wang2017multi} &
  98.6 &
  97.4 &
  96.3 &
  96.2 &
  75.2 &
  92.4 &
  96.5 &
  97.1 &
  76.5 &
  92.0 &
  87.7 &
  96.8 &
  97.5 &
  93.8 &
  98.5 &
  81.6 &
  93.7 &
  82.8 &
  98.6 &
  89.3 &
  91.9 \\
Atten-Reinforce* \cite{chen2018recurrent} &
  98.6 &
  97.1 &
  97.1 &
  95.5 &
  75.6 &
  92.8 &
  96.8 &
  97.3 &
  78.3 &
  92.2 &
  87.6 &
  96.9 &
  96.5 &
  93.6 &
  98.5 &
  81.6 &
  93.1 &
  83.2 &
  98.5 &
  89.3 &
  92.0 \\
RLSD \cite{zhang2018multilabel}  & 96.4 & 92.7 & 93.8 & 94.1 & 71.2 & 92.5 & 94.2 & 95.7 & 74.3 & 90.0 & 74.2 & 95.4 & 96.2 & 92.1 & 97.9 & 66.9 & 93.5 & 73.7 & 97.5 & 87.6 & 88.5 \\  
SSGRL \cite{chen2019learning}  & 99.5 & 97.1 & 97.6 & 97.8 & 82.6 & 94.8 & 96.7 & 98.1 & 78.0 & 97.0 & 85.6 & 97.8 & 98.3 & 96.4 & 98.1 & 84.9 & 96.5 & 79.8 & 98.4 & 92.8 & 93.4 \\
ML-GCN $\dagger$ \cite{chen2019multi} & 99.5 & 98.5 & 98.6& 98.1 & 80.8 & 94.6 & 97.2 & 98.2 & 82.3 & 95.7 & 86.4 & 98.2 & 98.4 & 96.7 & 99.0 & 84.7 & 96.7 & 84.3 & 98.9 & 93.7 & 94.0 \\
ADD-GCN $\dagger$ \cite{ye2020attention} & 99.7 & 98.5 & 97.6 & 98.4 & 80.6 & 94.1 & 96.6 & 98.1 & 80.4 & 94.9 & 85.7 & 97.9 & 97.9 & 96.4 & 99.0 & 80.2 & 97.3 & 85.3 & 98.9 & 94.1 & 93.6 \\
ASL-L $\dagger$ \cite{ridnik2021asymmetric} & \textbf{100.0} & 98.8 & 99.2 & 98.5 & \textbf{87.9} & 97.5 & \textbf{98.5} & 98.3 & 82.1 & 97.8 & 90.6 & 98.7 & \textbf{99.5 }& 97.9 & \textbf{99.3} & 89.9 & 99.5 & 84.1 & 99.4 & \textbf{96.7} & 95.8\\
TDRG $\dagger$ \cite{zhao2021transformer}  & 99.9 & 98.9 & 98.4 & 98.7 & 81.9 & 95.8 & 97.8& 98.0 & 85.2 & 95.6 & 89.5 & 98.8 & 98.6 & 97.1 & 99.1 & 86.2 & 97.7 & 87.2 & 99.1 & 95.3 & 95.0\\
CSRA $\dagger$ \cite{zhu2021residual} & 99.9 & 98.4 & 98.1 & \textbf{98.8} & 82.2 & 95.2 & 97.9 & 97.9 & 84.6 & 94.8 & 90.6 & 98.0 & 97.6 &96.3 & 99.1 &86.6 & 95.6 & 88.3 & 98.9 & 94.4
    & 94.7 
\\ \hline \hline

 

KMCL (TResNet-M) & 99.9&	98.7	&98.4	&98.0	&79.3	&97.3	&97.8	&97.3	&82.8	&\textbf{98.6}	&\textbf{91.1}	&98.1	&99.0	&97.7	&98.7	&\textbf{87.1}	&\textbf{99.6}	&\textbf{89.6}	&\textbf{99.7}	&96.0 & 95.2 \\
 
KMCL (TResNet-L) & \textbf{100.0} &	\textbf{99.4}	& \textbf{99.3}	& 98.5	& 84.5	&\textbf{98.0}	& 97.9	&\textbf{99.0}	&\textbf{86.2}	&\textbf{99.8}	&\textbf{90.7}	&\textbf{99.6}	&\textbf{99.5}	&\textbf{98.9}	&\textbf{99.3}	&85.8	&\textbf{99.9}	&\textbf{91.6}	&\textbf{99.6}	&95.8& \textbf{96.2} \\

\specialrule{2pt}{1pt}{1pt}
\vspace{-3em}
\end{tabular}
}
}
\end{table}

\vspace{-8mm}
\begin{figure}[htp]
\begin{minipage}{0.73\textwidth}
\centering
\tiny
%
\resizebox{\textwidth}{!}{
\setlength{\tabcolsep}{0.8mm}
\renewcommand{\arraystretch}{0.9}
\setlength\tabcolsep{3pt}
\begin{tabular}{c|c|c|c|c|c|c|c|c|c|c|c|c|c|c}
\specialrule{2pt}{1pt}{1pt}
\multirow{2}{*}{Methods} & \multirow{2}{*}{($R_{tr/ts}$)} & \multirow{2}{*}{mAP} & \multicolumn{6}{c|}{All}  & \multicolumn{6}{c}{Top 3}
\\
&        &  & CP   & CR   & CF1  & OP   & OR   & OF1  & CP   & CR   & CF1  & OP   & OR   & OF1  \\ \hline \hline
CNN-RNN \cite{wang2016cnn}              & $(224)$ & 61.2          & -    & -    & -    & -    & -    & -    & 66.0 & 55.6 & 60.4 & 69.2 & 66.4 & 67.8 \\
SRN \cite{zhu2017learning}                   & $(224)$  & 77.1                 & 81.6 & 65.4 & 71.2 & 82.7 & 69.9 & 75.8 & 85.2 & 58.8 & 67.4 & 87.4 & 62.5 & 72.9 \\
PLA  \cite{yazici2020orderless}            & $(288)$     & -                 & 80.4 & 68.9 & 74.2 & 81.5 & 73.3 & 77.2 & -    & -    & -    & -    & -    & -    \\
\textbf{KMCL-(TResNet-M)} & $(224)$ &   \textbf{82.1}  &  \textbf{84.1 } & \textbf{72.0} & \textbf{77.6} & \textbf{85.0} &  \textbf{76.1} & \textbf{79.7} & \textbf{87.9} & \textbf{64.0} &  \textbf{74.1} & \textbf{90.0}& \textbf{66.3} &\textbf{76.3 }    \\ \hline \hline
RNN-Attention \cite{wang2017multi}         &$(448)$   & -                    & -    & -    & -    & -    & -    & -    & 79.1 & 58.7 & 67.4 & 84.0 & 63.0 & 72.0 \\
ResNet-101 *\cite{he2016deep}              &$(448)$ & 78.6                 & 82.4 & 65.5 & 73.0 & 86.0 & 70.4 & 77.4 & 85.9 & 58.6 & 69.7 & 90.5 & 62.8 & 74.1 \\
ML-GCN \cite{chen2019multi}                 & $(448)$& 83.0                 & 85.1 & 72.0 & 78.0 & 85.8 & 75.4 & 80.3 & 89.2 & 64.1 & 74.6 & 90.5 & 66.5 & 76.7 \\
KSSNet \cite{wang2020multi}                 &$(448)$  & 83.7                 & 84.6 & 73.2 & 77.2 & 87.8 & 76.2 & 81.5 & -    & -    & -    & -    & -    & -    \\
TDRG \cite{zhao2021transformer}&  $(448)$  & 84.6  & 86.0 & 73.1 & 79.0 & 86.6 & 76.4 & 81.2 & 89.9 & 64.4 & 75.0 & 91.2 & 67.0 & 77.2 \\ 
ResNet-101 + CSRA \cite{zhu2021residual} &  $(448)$& 83.5 &84.1 &72.5 &77.9 &85.6 &75.7 &80.3 &88.5 &64.2 &74.4 &90.4 &66.4 &76.5 \\
ResNet-cut + CSRA \cite{zhu2021residual} &  $(448)$ &85.6 &86.2 &74.9 &80.1 &\textbf{86.6} &78.0 &82.1 &90.1 &65.7 &76.0 &91.4 &67.9 & 77.9 \\
ASL \cite{ridnik2021asymmetric} & $(448)$ & 88.4 & 85.0 & \textbf{81.9} & 83.4 & 85.2 & \textbf{84.2} & 84.7 & 90.0 & 69.8 & 78.6 & 91.6 & \textbf{71.3} & 80.2\\
\textbf{KMCL-(TResNet-L)} & $(448)$ &  \textbf{88.6}   &  \textbf{87.7}  & 81.6 & \textbf{83.6} & 86.3 & 83.6 & \textbf{84.9} & \textbf{90.6}& \textbf{70.0 }& \textbf{78.9} &\textbf{92.1} & 71.2 &\textbf{80.3 }\\
\specialrule{2pt}{1pt}{1pt}
\end{tabular}
}
\end{minipage}
\hfill
\begin{minipage}{0.25\textwidth}
    \centering
    \includegraphics[width=\textwidth]{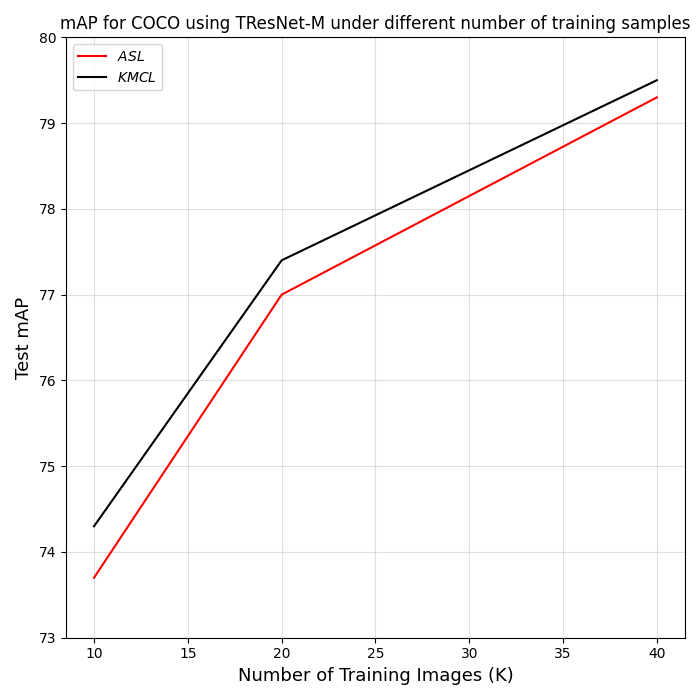}
  \end{minipage}
  \vspace*{-0.5mm}
\caption{(Left) Comparisons with state-of-the-art methods on MS-COCO \cite{lin2014microsoft}, sorted by input image resolution for training and testing. (Right) Comparing KMCL and ASL on MS-COCO for varying numbers of training samples}
\label{fig:mscoco}
\vspace{-1em}
\end{figure}

\begin{table}[htp]
\tiny
\centering
\setlength\tabcolsep{3pt}
\vspace*{-6mm}
	\caption{Comparisons with state-of-the-art methods on the Chest-Xray14 dataset.}
	\label{table: Xray benchmark}
	\resizebox{\columnwidth}{!}{
\begin{tabular}{c|c|c|c|c|c|c|c||c|c|c|c|c|c|c|c|c|c|c|c|c|c}
\specialrule{2pt}{1pt}{1pt}
\multirow{2}{2em}{Method} & \multicolumn{7}{c|}{Overall Performance} & \multicolumn{14}{c}{Class-wise AUC Score} \\
\cline{2-22}
     & mAP & OP & OR & OF1 & CP & CR & CF1
& At. & Ca. & Ef. & In. & M. & N. & Pm. & Pt. & Co. & Ed. & Em. & Fi. & Plet. & He. \\ \specialrule{2pt}{1pt}{1pt}
    ML-GCN (Binary) \cite{chen2019multi} & 26.3 & \textbf{53.1} & 23.2 & 32.3  & \textbf{35.8} & 10.6 & 16.4
    & 73.1& 81.5& 71.7& 67.3& 51.6& 69.9& 59.1& 76.9& 60.9& 78.5& 82.2& \textbf{76.7}& 67.5& \textbf{94.6} \\  
    
    ASL (TResNet-L) \cite{ridnik2021asymmetric} & 25.5 & 47.1 & 22.8 & 30.7  & 33.8 & 13.8 & 19.7
    &72.0& 81.1& 71.0& 63.6& 49.6& 72.0& 57.6& 78.4& 64.7& 78.8& 80.1& 75.4& 66.9& 88.9 \\  
    
    TDRG  \cite{zhao2021transformer}& 24.2  & 51.0 & 28.4 & 36.5 & 21.2 & 13.5 & 16.5 
    & 71.4& 82.1& 75.5& 67.2& 49.2& 70.2& 61.4& 73.9& \textbf{65.0}& 76.9& 76.3& 69.8& 68.5& 83.9  \\ 
    
    CSRA \cite{zhu2021residual}&  26.3 & 52.0 & 26.2 & 34.8 & 24.6 & 12.0 & 16.1 
    & 71.5& 83.2& 76.4& 66.9& 56.8& 74.7& 59.7&\textbf{ 81.8}& 62.7& 77.6& 86.3& 73.2& 67.1& 90.9 \\ 
    \hline
    KMCL (TResNet-M) & \textbf{30.0} & 51.1 & \textbf{35.4} & \textbf{40.9}  & 33.1 & \textbf{18.8} & \textbf{23.7}
    & \textbf{73.1}& \textbf{88.8}& \textbf{77.8}& 67.0& \textbf{59.7}& \textbf{75.9}& 59.2& \textbf{81.8}& 63.5& \textbf{80.4}& 85.5& 75.1& \textbf{70.1}& 93.3 \\ 
    
    KMCL (TResNet-L) & \textbf{31.5} & 47.9 & \textbf{41.2 }& \textbf{43.4 } & 33.1 & \textbf{25.4} & \textbf{28.8}
    & \textbf{73.7}& \textbf{88.4}& \textbf{77.9}& \textbf{67.8}& 53.1& \textbf{75.3}& \textbf{61.7}& \textbf{82.0}& 61.8& \textbf{80.1}& \textbf{91.5}& 74.6& \textbf{69.7}& 92.9 \\ 
\specialrule{2pt}{1pt}{1pt}
\vspace{-2.7em}
\end{tabular}
}
\end{table}\vspace{-1.0em}

\textbf{How well KMCL generalizes to medical imaging datasets?}

\begin{wraptable}[8]{r}{7cm}
\tiny
\centering
\vspace{-1.4em}
\setlength\tabcolsep{3pt}
	\caption{Comparisons with state-of-the-art methods on the ADP dataset.}
	\label{table: ADP benchmark}
	\resizebox{\columnwidth}{!}{
\begin{tabular}{c|c|c|c|c|c|c|c|c|c}

\specialrule{2pt}{1pt}{1pt}
 \multirow{2}{2em}{Method} & \multicolumn{7}{c|}{Performance} & \multicolumn{2}{c}{Complexity} \\
\cline{2-10}
    
     & mAP & OP & OR & OF1 & CP & CR & CF1 & Parameters (MM) & GMAC
\\ \specialrule{2pt}{1pt}{1pt}
ML-GCN (Binary) \cite{chen2019multi} & 94.9 & 92.0 & 86.9 & 89.7  & 91.8& 87.0& 89.3
&44.90 & 31.39\\ 
    
    ASL (TResNet-L) \cite{ridnik2021asymmetric} & 96.1 & 92.1 & 90.7 & 91.4  & 92.5 & 89.2 & 90.8
    &44.14 & 35.28\\ 
    
    TDRG  \cite{zhao2021transformer}& 95.5 & \textbf{94.3} & 86.2 & 90.5  & 94.6 & 84.8 & 89.4
    &75.20 & 64.40\\ 
    
    CSRA \cite{zhu2021residual} & 96.1 & 93.0& 89.7 & 91.7 & 93.1& 88.6 & 90.8
    &42.52 & 31.39\\ 
    
    \hline
    KMCL (TResNet-M) & 95.1 & 94.2 & \textbf{91.0 }& 90.4 & \textbf{94.7} & 88.9 & 89.8 
    &29.41 & 5.74\\ 
    
    KMCL (TResNet-L) & \textbf{96.5}  & 92.7 & \textbf{92.9} & \textbf{92.8} & 92.6 & \textbf{92.0} &  \textbf{92.3}
    &44.20 & 35.28\\ 
\specialrule{2pt}{1pt}{1pt}
\vspace{-2.8em}
\end{tabular}
}
\end{wraptable}

\vspace{-0.5em}
We evaluate KMCL against SOTA methods on medical imaging datasets presented in Tables \ref{table: Xray benchmark} and\ref{table: ADP benchmark}. The recall is a crucial factor in these datasets, as it reflects the likelihood of missing a medical diagnosis. The proposed method achieves a superior tradeoff between precision and recall by significantly improving recall metrics while maintaining competitive precision scores, including SOTA mAP. On the ADP dataset, KMCL outperforms the surveyed SOTA with margins of $0.4\%$, $2.2\%$, and $2.8\%$ for mAP, OR, and CR, respectively. Similarly, on the ChestX-ray14 dataset, both TResNet-M and TResNet-L models exhibit significant improvements, with our best model surpassing SOTA results by $5.2\%$, $7.0\%$, and $11.6\%$ in mAP, OR, and CR, respectively. In comparison, competing methods such as ML-GCN \cite{chen2019multi} use label correlation but suffer from increased computational complexity and a multi-stage approach, as shown in Table \ref{table: ADP benchmark}. However, our method surpasses the SOTA while maintaining a small model size and low GMAC scores. These findings highlight the advantage of KMCL in computationally constrained environments.

\textbf{How KMCL's performance varies with different similarity measurements?} In this ablation study, we examine the impact of changing the Battacharya coefficient to either Mahalanobis kernel similarity or Gaussian kernel similarity in the KMCL framework (Corrolary \ref{corr1} (i) and (ii)). Under the Mahalanobis kernel similarity, the performance decreases across the PascalVOC and ADP, as indicated in Table \ref{table: Abaltive Cases}. This is likely due to the constraint that the variance must be identical across all classes, leading to an inability to capture entropy and uncertainty as reported in Section \ref{sec_Bhattacharyya}.

\begin{wraptable}[7]{r}{7cm}
\vspace{-0.00001em}
    \belowcaptionskip    
\setlength\tabcolsep{}
	\caption{Ablative comparison for similarity measures and kernel representation cases.}
	\label{table: Abaltive Cases}
	\resizebox{\columnwidth}{!}{
\begin{tabular}{c|c||c|c|c|c|c|c|c|c|c|c}
\specialrule{2pt}{1pt}{1pt}
 \multicolumn{2}{c||}{Configuration} & \multicolumn{7}{c|}{ADP} & PascalVOC & \multicolumn{2}{c}{Complexity} \\
\hline
    Similarity Metric & Case & mAP & OP & OR & OF1 & CP & CR & CF1 & mAP & Params(MM) & GMAC
\\ \specialrule{2pt}{1pt}{1pt}
Bhattacharyya & Anisotropic  & \textbf{95.4} & 94.0 & \textbf{92.7} & 90.6  & \textbf{94.8} & \textbf{90.7} & \textbf{90.5} & \textbf{95.4}
& 104.91 & 5.81\\
Bhattacharyya & Isotropic  & 95.1 & \textbf{94.2} & 91.0& 90.4 & 94.7 & 88.9 & 89.8 &  95.2
& 29.41 & \textbf{5.74}\\
Mahalanobis & -  & 94.7 & 92.0 & 92.4 & \textbf{90.9 } & 92.6 & 90.5 & 90.4 & 95.1
& 71.34 & 5.78\\
Gaussian Kernel & -  & 94.5 & 91.5 & 89.7 & 90.6  & 92.3 & 86.5 & 89.3 & 95.0
& \textbf{29.40} &\textbf{ 5.74}\\ 
\specialrule{2pt}{1pt}{1pt}
\vspace{-3em}
\end{tabular}
}
\end{wraptable}
  Similarly, when utilizing Gaussian kernel similarity, the performance further deteriorates because the model is constrained to learn a single variance value that applies to both the label classes and feature dimensions. Therefore, it is more meaningful to use the Bhattacharyya coefficient since it evaluates the generalized variances of the kernels and identifies similarities in their orientation, shape, and means (Eq. \ref{eq:BDGen}). We further investigate the assumptions from both isotropic and anisotropic cases of the exponential kernel representations in KMCL framework as discussed in Corrolary \ref{cor2}. While the anisotropic case leads to an improved performance as shown in Table \ref{table: Abaltive Cases}, but results in an increase in learnable parameters at the cost of higher computational complexity. By incorporating variances over the feature dimension, we better capture epistemic uncertainty and achieve enhanced overall results. Thus, if computational resources are available, one could best leverage our framework in the anisotropic case to achieve sota results.

\label{sec:vis}
\begin{wrapfigure}[11]{r}{0.7\textwidth}
\vspace{-1.6em}
    \centering
    \includegraphics[width=\textwidth]{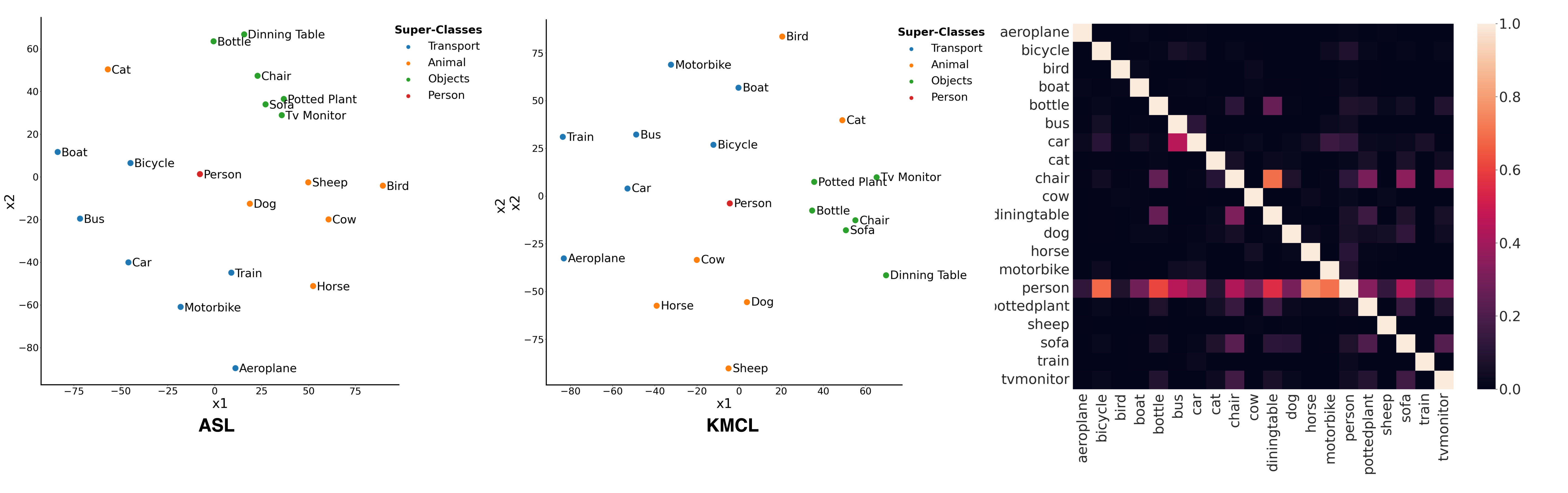}
    \vspace{-2em}
    \caption{Reduced t-SNEs for ASL (left) and KMCL(Center) on PascalVOC color-coded by user-defined super-classes in the legend; (Right) ground truth correlation matrix for PascalVOC.}
    \label{fig:TSNEs}
\end{wrapfigure}

\textbf{Intuitive Visualizations.} 
KMCL presents an end-to-end framework for contrastive learning that has achieved quantitatively significant results compared to existing methods. In this section, we visualize how the learned feature representation incorporates label correlation and epistemic uncertainty. Figure \ref{fig:TSNEs} shows a reduced t-SNE \cite{van2008visualizing} visualization of the feature representation for ASL and KMCL on the Pascal VOC dataset. Both methods accurately discriminate between different classes, as seen from the plotted centroids of each cluster. Notably, both methods exhibit a clustering pattern based on user-defined super-classes (e.g., \textit{car} and \textit{bus} are both forms of Transportation). Upon analyzing the ground truth correlation matrix, it becomes apparent that KMCL captures label correlation more effectively. Specifically, the \textit{sofa} class exhibits the highest correlation with the \textit{chair} class, resulting in their closer proximity in the t-SNE visualization for KMCL compared to ASL.

\begin{wrapfigure}[8]{r}{0.65\textwidth}
 \vspace{-2em}
    \centering
    \includegraphics[width=\textwidth]{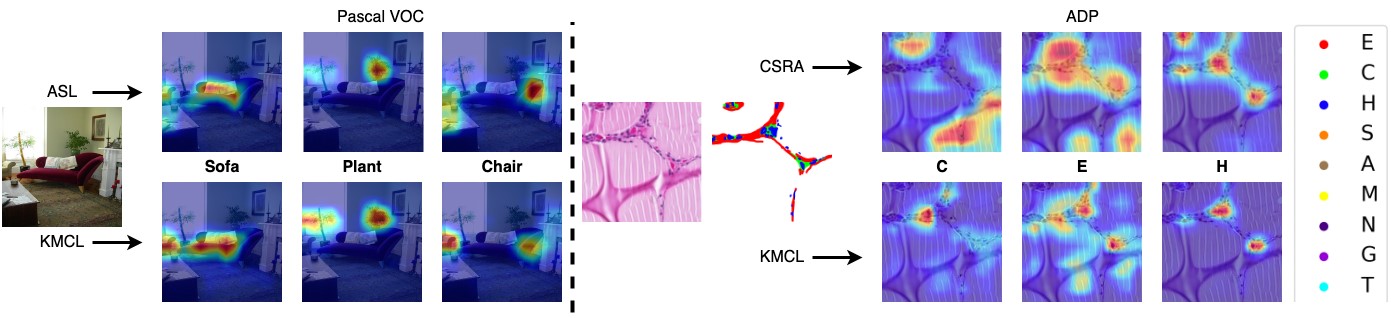}
    \vspace{-2em}
    \caption{GradCam visualization of KMCL and competing SOTA method. Bolded class labels indicate instances where KMCL outperforms SOTA by a large margin.}
    \label{fig:GradCam}
\end{wrapfigure}
Figure \ref{fig:GradCam} showcases the GradCam visualization for KMCL and a competing SOTA method. KMCL effectively distinguishes the \textit{sofa} and \textit{chair} classes, consistent with the t-SNE visualization results. Moreover, by capturing epistemic uncertainty from the kernel representation, our method accurately identifies the correct classes in the ADP sample with minimal extraneous activations. For more visualizations, please refer to the Supplementary material. 

\section*{Broader Impact}
KMCL provides an end-to-end supervised contrastive learning framework for multilabel datasets. It requires fewer resources for the design and implementation of downstream tasks such as classification. Contrastive learning methods like \cite{chen2020simple, khosla2020supervised, malkinski2020multi} typically involve two stages of encoder training and fine-tuning for the task, which can take several hundred epochs. In contrast, KMCL only requires one stage of training with significantly fewer epochs. This translates into a much smaller carbon emission footprint, as highlighted in \cite{fu2021reconsidering} for using more compact models for training. Although KMCL has been successfully applied in computer vision and medical imaging domains, its effectiveness has not yet been tested for segmentation/detection tasks or in other modalities like natural language processing. In future work, we will consider broadening our experiments for further validation. Additionally, we believe that society can benefit from the theoretical analysis of the similarity metrics presented in this paper, which can be adapted to different application domains.



\section*{Acknowledgment}
Authors would like to thank Rahavi Selvarajan, Xiao Hu and Jiarui Zhang for their assistant and helpful discussion.

{\small
\bibliographystyle{ieee_fullname}
\bibliography{egbib}
}

\appendix

\appendix
\section{Appendix} \label{sec:appendix}
\subsection{Proof of Remark 1.} \label{sec:appendix}
\begin{proof}
The Bhattacharyya coefficient between the normalized $p(\mathbf{x}):= K_{\boldsymbol{\Sigma}_{p}}(\mathbf{x}, \boldsymbol{\mu}_{p}) = \exp{\left(-\frac{1}{2}\|\mathbf{x} - \boldsymbol{\mu}_{p}\|^{2}_{\boldsymbol{\Sigma}_{p}^{-1}}\right)}$ and $q(\mathbf{x}):= K_{\boldsymbol{\Sigma}_{q}}(\mathbf{x}, \boldsymbol{\mu}_{q}) = \exp{\left(-\frac{1}{2}\|\mathbf{x} - \boldsymbol{\mu}_{q}\|^{2}_{\boldsymbol{\Sigma}_{q}^{-1}}\right)}$ is defined as 
\begin{equation} 
    \rho\big(p(\mathbf{x}), q(\mathbf{x}) \big)  = \int_{\mathcal{X}}\Big(\dfrac{p(\mathbf{x})}{\int_{\mathcal{X}} p(\mathbf{x}) d\mathbf{x}}\Big)^{\frac{1}{2}}\Big(\dfrac{q(\mathbf{x})}{\int_{\mathcal{X}} q(\mathbf{x}) d\mathbf{x}}\Big)^{\frac{1}{2}}d\mathbf{x} =  \frac{\int_{\mathcal{X}}p(\mathbf{x})^{\frac{1}{2}}q(\mathbf{x})^{\frac{1}{2}}d\mathbf{x}}{\sqrt{\int_{\mathcal{X}}p(\mathbf{x}) d\mathbf{x}}\sqrt{\int_{\mathcal{X}}q(\mathbf{x}) d\mathbf{x}}}. \nonumber 
\end{equation}
To begin, we expand the integrand part of the enumerator, i.e., $\sqrt{p(\mathbf{x})q(\mathbf{x})}$ as follows:
\begin{flalign} \label{eq:1}
\exp{\Big(-\dfrac{1}{4}\mathbf{x}^{T}\big(\boldsymbol{\Sigma}_{p}^{-1}+\boldsymbol{\Sigma}_{q}^{-1}\big)\mathbf{x}+\dfrac{1}{2}\big(\boldsymbol{\Sigma}_{p}^{-1}\boldsymbol{\mu}_{p}+\boldsymbol{\Sigma}_{q}^{-1}\boldsymbol{\mu}_{q}}\big)^{T}\mathbf{x} -\dfrac{1}{4}\big(\boldsymbol{\mu}_{p}^{T}\boldsymbol{\Sigma}_{p}^{-1}\boldsymbol{\mu}_{p} + \boldsymbol{\mu}_{q}^{T}\boldsymbol{\Sigma}_{q}^{-1}\boldsymbol{\mu}_{q} \big)\Big). 
\end{flalign}
In order to overcome the challenge of integrating the derived integrand in Equation \ref{eq:1}, we will introduce a new approach. We will represent $\sqrt{p(\mathbf{x})q(\mathbf{x})}$ as the product of a constant value, denoted as $h(\boldsymbol{\mu}_{p}, \boldsymbol{\mu}_{q}, \boldsymbol{\Sigma}_{p}, \boldsymbol{\Sigma}_{q})$, and a newly defined anisotropic multivariate squared exponential kernels, denoted as $r(\mathbf{x}):= K_{\boldsymbol{\Sigma}_{r}}(\mathbf{x}, \boldsymbol{\mu}_{r})$. This formal representation can be expressed as follows:
\begin{flalign} \label{eq:2}
\sqrt{p(\mathbf{x})q(\mathbf{x})} = h(\boldsymbol{\mu}_{p}, \boldsymbol{\mu}_{q}, \boldsymbol{\Sigma}_{p}, \boldsymbol{\Sigma}_{q})r(\mathbf{x}). 
\end{flalign}
We defined the new exponential kernel of Equation \ref{eq:2} as
\begin{flalign} \label{eq:3}
r(\mathbf{x}) := K_{\boldsymbol{\Sigma}_{r}}(\mathbf{x}, \boldsymbol{\mu}_{r}) = \exp{\left(-\frac{1}{2}\|\mathbf{x} - \boldsymbol{\mu}_{r}\|^{2}_{\boldsymbol{\Sigma}_{r}^{-1}}\right)} = \exp{\big(-\dfrac{1}{2}(\mathbf{x}-\boldsymbol{\mu}_{r})^{T}\boldsymbol{\Sigma}_{r}^{-1}(\mathbf{x}-\boldsymbol{\mu}_{r})\big)},
\end{flalign}
where $\boldsymbol{\Sigma}_{r} \triangleq \big(\dfrac{1}{2}\boldsymbol{\Sigma}_{p}^{-1}+\dfrac{1}{2}\boldsymbol{\Sigma}_{q}^{-1}\big)^{-1}$ and $\boldsymbol{\mu}_{r} \triangleq \boldsymbol{\Sigma}_{p}\big(\dfrac{1}{2}\boldsymbol{\Sigma}_{p}^{-1}\boldsymbol{\mu}_{p}+\dfrac{1}{2}\boldsymbol{\Sigma}_{q}^{-1}\boldsymbol{\mu}_{q}\big)$. Once the values of $\boldsymbol{\Sigma}_{r}$ and $\boldsymbol{\mu}_{r}$ are replaced in Equation \ref{eq:3}, the kernel $r(\mathbf{x})$ will be
\begin{flalign} \label{eq:4}
r(\mathbf{x})  = &\exp{\Big(-\dfrac{1}{4}\mathbf{x}^{T}\big(\boldsymbol{\Sigma}_{p}^{-1}+\boldsymbol{\Sigma}_{q}^{-1}\big)\mathbf{x}  + \dfrac{1}{2}\big(\boldsymbol{\Sigma}_{p}^{-1}\boldsymbol{\mu}_{p}+\boldsymbol{\Sigma}_{q}^{-1}\boldsymbol{\mu}_{p}}\big)^{T}\mathbf{x} -\dfrac{1}{4}\big(\boldsymbol{\Sigma}_{p}^{-1}\boldsymbol{\mu}_{p}+\boldsymbol{\Sigma}_{q}^{-1}\boldsymbol{\mu}_{p}\big)^{T} \nonumber \\ 
&\hspace*{4.3cm}+\big(\boldsymbol{\Sigma}_{p}^{-1}+\boldsymbol{\Sigma}_{q}^{-1}\big)^{-1}\big(\boldsymbol{\Sigma}_{p}^{-1}\boldsymbol{\mu}_{p}+\boldsymbol{\Sigma}_{q}^{-1}\boldsymbol{\mu}_{p}\big)\Big).
\end{flalign}
By substituting Equations \ref{eq:1} and \ref{eq:4} into Equation \ref{eq:2}, we obtain the closed-form expression of $h(\boldsymbol{\mu}_{p}, \boldsymbol{\mu}_{q}, \boldsymbol{\Sigma}_{p}, \boldsymbol{\Sigma}_{q})$ as presented below.
\begin{flalign} \label{eq:5}
\exp{\Big(\frac{-1}{4}(
\boldsymbol{\mu}_{p}^{T}\big(\boldsymbol{\Sigma}_{p}^{-1}-\boldsymbol{\Sigma}_{p}^{-1}(\boldsymbol{\Sigma}_{p}^{-1}+\boldsymbol{\Sigma}_{q}^{-1})^{-1}\boldsymbol{\Sigma}_{p}^{-1}\big)\boldsymbol{\mu}_{p}+
\boldsymbol{\mu}_{q}^{T}\big(\boldsymbol{\Sigma}_{q}^{-1}-\boldsymbol{\Sigma}_{q}^{-1}(\boldsymbol{\Sigma}_{p}^{-1}+\boldsymbol{\Sigma}_{q}^{-1})^{-1}\boldsymbol{\Sigma}_{q}^{-1}\big)\boldsymbol{\mu}_{q}}
\nonumber  \\
-\boldsymbol{\mu}_{p}^{T}\big(\boldsymbol{\Sigma}_{p}^{-1}(\boldsymbol{\Sigma}_{p}^{-1}+\boldsymbol{\Sigma}_{q}^{-1})^{-1}\boldsymbol{\Sigma}_{q}^{-1}\big)\boldsymbol{\mu}_{q}
-\boldsymbol{\mu}_{q}^{T}\big(\boldsymbol{\Sigma}_{q}^{-1}(\boldsymbol{\Sigma}_{p}^{-1}+\boldsymbol{\Sigma}_{q}^{-1})^{-1}\boldsymbol{\Sigma}_{p}^{-1}\big)\boldsymbol{\mu}_{p}
)\Big)
\end{flalign}
Given the fact that $\boldsymbol{\Sigma}_{p}^{-1}-\boldsymbol{\Sigma}_{p}^{-1}(\boldsymbol{\Sigma}_{p}^{-1}+\boldsymbol{\Sigma}_{q}^{-1})^{-1}\boldsymbol{\Sigma}_{p}^{-1} = \boldsymbol{\Sigma}_{q}^{-1}-\boldsymbol{\Sigma}_{q}^{-1}(\boldsymbol{\Sigma}_{p}^{-1}+\boldsymbol{\Sigma}_{q}^{-1})^{-1}\boldsymbol{\Sigma}_{q}^{-1} = \boldsymbol{\Sigma}_{p}^{-1}(\boldsymbol{\Sigma}_{p}^{-1}+\boldsymbol{\Sigma}_{q}^{-1})^{-1}\boldsymbol{\Sigma}_{q}^{-1} = \boldsymbol{\Sigma}_{q}^{-1}(\boldsymbol{\Sigma}_{p}^{-1}+\boldsymbol{\Sigma}_{q}^{-1})^{-1}\boldsymbol{\Sigma}_{p}^{-1} = (\boldsymbol{\Sigma}_{p}+\boldsymbol{\Sigma}_{q})^{-1}$ \cite{woodbury1950inverting}, we can simplify Equation \ref{eq:5} and derive
\begin{flalign}
 & \exp{\big(\frac{-1}{4}\boldsymbol{\mu}_{p}^{T}\big(\boldsymbol{\Sigma}_{p}+\boldsymbol{\Sigma}_{q}\big)^{-1}\boldsymbol{\mu}_{p}+\boldsymbol{\mu}_{q}^{T}\big(\boldsymbol{\Sigma}_{p}+\boldsymbol{\Sigma}_{q}\big)^{-1}\boldsymbol{\mu}_{q}-\boldsymbol{\mu}_{p}^{T}\big(\boldsymbol{\Sigma}_{p}+\boldsymbol{\Sigma}_{q}\big)^{-1}\boldsymbol{\mu}_{q}-\boldsymbol{\mu}_{q}^{T}\big(\boldsymbol{\Sigma}_{p}+\boldsymbol{\Sigma}_{q}\big)^{-1}\boldsymbol{\mu}_{p}\big)}, \nonumber
\end{flalign}
where can be further simplified to yield the following expression:
\begin{flalign} \label{eq:6}
h(\boldsymbol{\mu}_{p}, \boldsymbol{\mu}_{q}, \boldsymbol{\Sigma}_{p}, \boldsymbol{\Sigma}_{q}) = \exp{\big(-\dfrac{1}{8}(\boldsymbol{\mu}_{p}-\boldsymbol{\mu}_{q}})^{T}\boldsymbol{\Sigma}^{-1}(\boldsymbol{\mu}_{p}-\boldsymbol{\mu}_{q})\big),
\end{flalign}
where $\boldsymbol{\Sigma} 
 = \frac{\boldsymbol{\Sigma}_{p}+\boldsymbol{\Sigma}_{q}}{2}$. Ultimately, by utilizing the definition of the Bhattacharyya coefficient, Equation \ref{eq:2}, and Equation \ref{eq:6}, we can deduce the following conclusion:
 \begin{flalign} \label{eq:8}
\rho\big(p(\mathbf{x}), q(\mathbf{x})\big) &=  \frac{\int_{\mathbb{R}^{M}}p(\mathbf{x})^{\frac{1}{2}}q(\mathbf{x})^{\frac{1}{2}}d\mathbf{x}}{\sqrt{\int_{\mathbb{R}^{M}}p(\mathbf{x}) d\mathbf{x}}\sqrt{\int_{\mathbb{R}^{M}}q(\mathbf{x}) d\mathbf{x}}} =\frac{\int_{\mathbb{R}^{M}}h(\boldsymbol{\mu}_{p}, \boldsymbol{\mu}_{q}, \boldsymbol{\Sigma}_{p}, \boldsymbol{\Sigma}_{q})r(\mathbf{x})d\mathbf{x}}{\sqrt{\int_{\mathbb{R}^{M}}p(\mathbf{x}) d\mathbf{x}}\sqrt{\int_{\mathbb{R}^{M}}q(\mathbf{x}) d\mathbf{x}}}  \nonumber \\ 
& = \frac{h(\boldsymbol{\mu}_{p}, \boldsymbol{\mu}_{q}, \boldsymbol{\Sigma}_{p}, \boldsymbol{\Sigma}_{q}) \int_{\mathbb{R}^{M}}\big|2\pi \boldsymbol{\Sigma}_{r}\big|^{\frac{1}{2}}\mathcal{N}(\mathbf{x};\boldsymbol{\mu_{r}}, \boldsymbol{\Sigma_{r}})d\mathbf{x}}{\sqrt{\int_{\mathbb{R}^{M}}\big|2\pi \boldsymbol{\Sigma}_{p}\big|^{\frac{1}{2}}\mathcal{N}(\mathbf{x};\boldsymbol{\mu_{p}}, \boldsymbol{\Sigma_{p}}) d\mathbf{x}}\sqrt{\int_{\mathbb{R}^{M}}\big|2\pi \boldsymbol{\Sigma}_{q}\big|^{\frac{1}{2}}\mathcal{N}(\mathbf{x};\boldsymbol{\mu_{q}}, \boldsymbol{\Sigma_{q}}) d\mathbf{x}}} \nonumber \\
& = \frac{\big|\boldsymbol{\Sigma}_{r}\big|^{\frac{1}{2}}}{\big|\boldsymbol{\Sigma}_{p}\big|^{\frac{1}{4}}\big|\boldsymbol{\Sigma}_{q}\big|^{\frac{1}{4}}}h(\boldsymbol{\mu}_{p}, \boldsymbol{\mu}_{q}, \boldsymbol{\Sigma}_{p}, \boldsymbol{\Sigma}_{q}) = \frac{\big|2\boldsymbol{\Sigma}_{p}(\boldsymbol{\Sigma}_{p}+\boldsymbol{\Sigma}_{q})^{-1}\boldsymbol{\Sigma}_{q}\big|^{\frac{1}{2}}}{\big|\boldsymbol{\Sigma}_{p}\big|^{\frac{1}{4}}\big|\boldsymbol{\Sigma}_{q}\big|^{\frac{1}{4}}}h(\boldsymbol{\mu}_{p}, \boldsymbol{\mu}_{q}, \boldsymbol{\Sigma}_{p}, \boldsymbol{\Sigma}_{q}) \nonumber \\
& \overset{(a)}{=} \dfrac{\big|\boldsymbol{\Sigma}_{p}\big|^{\frac{1}{2}}\big|\boldsymbol{\Sigma}_{q}\big|^{\frac{1}{2}}}{\big|\boldsymbol{\Sigma}\big|^{\frac{1}{2}}}\exp{\big(-\dfrac{1}{8}(\boldsymbol{\mu}_{p}-\boldsymbol{\mu}_{q}})^{T}\boldsymbol{\Sigma}^{-1}(\boldsymbol{\mu}_{p}-\boldsymbol{\mu}_{q})\big),
\end{flalign}
where, $\boldsymbol{\Sigma} 
 = \frac{\boldsymbol{\Sigma}_{p}+\boldsymbol{\Sigma}_{q}}{2}$ and $(a)$ is followed by the probability property that the total area underneath a probability density function is $1$. The notation $\mathcal{N}(\mathbf{x};\boldsymbol{\mu}, \boldsymbol{\Sigma})$ represents a multivariate Gaussian probability distribution in M dimensions, characterized by a mean vector $\boldsymbol{\mu}$ and a covariance matrix $\boldsymbol{\Sigma}$. This completes the proof of Remark 1.
\end{proof}

\subsection{Forward Propagation in KMM.} \label{appendix:forward}
The KMM (Kernel Mixture Module) takes the feature vector $\mathbf{f}^{n} \in \mathbb{R}^{M}$ as input from the encoder network and produces the parameters for each exponential kernel component in the kernel mixture model. This transformation converts the feature vector into $3K$ values, where each $K$ represents the parameters for the $k$th kernel component (existing class), such as ${\mu_{k}^{n} \in \mathbb{R}, \sigma_{k}^{n} \in \mathbb{R}^{+}, \pi_{k}^{n} \in [0, 1]}$. The adaptive parameters are computed through forward propagation, employing suitable activation functions to ensure that the parameters satisfy their respective constraints. The activations corresponding to the parameters of the $k$th component for the KMM $\big(({a}_{k}^{{\mu}})^{n},({a}_{k}^{\sigma^{2}})^{n}, (a_{k}^{\pi})^{n}\big) $ are used to accomplish this, and they are calculated through the forward propagation of a fully connected layer by
\begin{flalign}  \label{eq:10}
({a}_{k}^{\mu})^{n} = \mathbf{w}_{k}^{\mu}\mathbf{f}^{n} + {b}^{\mu}_{k},  \hspace{1.5cm} ({a}_{k}^{\sigma^{2}})^{n} = \mathbf{w}_{k}^{\sigma^{2}}\mathbf{f}^{n} + {b}^{\sigma^{2}}_{k},  \hspace{1.5cm} (a_{k}^{\pi})^{n} = \mathbf{w}_{k}^{\pi}\mathbf{f}^{n} + b^{\pi}_k,
\end{flalign}
where, $\{\mathbf{w}_{k}^{\mu}, \mathbf{w}_{k}^{\sigma^{2}}, \mathbf{w}_{k}^{\pi}\} \in \mathbb{R}^{M}$ are the weights, and $\{{b}^{\mu}_{k}, {b}^{\sigma^{2}}_{k}, b^{\pi}_{k}\} \in \mathbb{R}$ represent the biases associated with $\{({a}_{k}^{\mu})^{n}, ({a}_{k}^{\sigma^{2}})^{n}, (a_{k}^{\pi})^{n}\}$, respectively. We make a minor revision to the idea of using nonlinear activation from \cite{variani2015gaussian, bishop2006pattern} by replacing softmax with sigmoid to normalize the mixture of coefficients and address multilabel issues. In the following, we define the nonlinear and linear transformations applied to ${({a}{k}^{\mu})^{n}, ({a}{k}^{\sigma^{2}})^{n}, (a_{k}^{\pi})^{n}}$ using
\begin{flalign} \label{eq:11}
    \pi_{k}^{n} = \dfrac{1}{1+\exp{(-(a_{k}^{\pi})^{n})}}, \hspace{1.25cm}
       {\mu}_{k}^{n} = ({a}_{k}^{\mu})^{n}, \hspace{1.25cm} (\sigma_{k}^{n})^{2} = ELU((a_{k}^{\sigma^{2}})^{n})+2+\epsilon,
\end{flalign}
where ELU($\cdot$) and $\epsilon$ are the exponential linear unit function \cite{clevert2015fast} and the hyperparameter used to ensure training stability, respectively. We use a modified ELU function rather than the exponential function as the activation on $(a_{k}^{\sigma^{2}})^{n}$ in order to ensure that variances remain non-negative ($(\sigma_{k}^{n})^{2} \geq 0$). This modification is necessary because the vanilla exponential function exhibits rapid growth for larger values, which can lead to training instability, particularly when dealing with high-variance datasets. It is important to note that there is no constraint on the mean ${\mu}_{k}^{n}$, as it is obtained directly from the activation $({a}_{k}^{\mu})^{n}$.

\section{Datasets}\label{sec:Datasets}

\begin{table}[h]

    \centering
    \resizebox{0.8\columnwidth}{!}{
    \small
    \begin{tabular}{c|c|c|c} 
    \specialrule{2pt}{1pt}{1pt}
    Dataset & \# of Images & \# of Classes & \# of Channels  \\
    \specialrule{2pt}{1pt}{1pt}
    PASCAL-VOC (2007) \cite{everingham2010pascal}& 9,963 & 20 & 3 \\ 
    MS-COCO \cite{lin2014microsoft}& 82,081 & 80 & 3\\ 
    ADP-L1  \cite{hosseini2019atlas} & 14,134 & 9& 3\\ 
    ChestXray-14 \cite{wang2017chestx}& 112,120
    & 14 & 1 \\ 
    \specialrule{2pt}{1pt}{1pt}
    \end{tabular}
    }
    \caption{Dataset Distribution Table}
    \label{tab:my_label}
\end{table}

\paragraph{PASCAL-VOC}
The PASCAL Visual Object Classes Challenge (2007) \cite{everingham2010pascal} is a common computer vision dataset used in multi-label classification. It contains a total of $9963$ images over $20$ classes, including 'cat', 'bottle', and 'person'. Being consistent with the state of the art, we trained our architecture on the \textit{trainval} set and evaluated it on the test set with a total of $5011$ and $4952$ images in each set, respectively. Referencing the relative frequency in the main paper, we can see that the number of classes per image to the total number of classes is heavily unbalanced, with the majority of images having only $2-4$ classes.

\paragraph{MS-COCO}
The Microsoft COCO dataset \cite{lin2014microsoft} is another common computer vision dataset used in multi-label classification. This dataset includes 82,081 training and 40,504 validation images across 80 different classes including 'person', 'bicycle', and 'elephant'. Following the state of the art, we test our method on the validation dataset making it comparable with competitive approaches.


\paragraph{ADP}

The Atlas of Digital Pathology for Hisotological Tissue Type Classification \cite{hosseini2019atlas} is composed of digital histology images taken from several organ tissues, including the colon, brain, stomach, etc. These images were generated via a Whole Slide Image (WSI) scanner. This database includes $17,668$ image patches that are multilabel in nature. The training, validation, and test sets contain $14,134$, $1767$, and $1767$ images respectively. This labeling scheme follows a three-tier hierarchy: L1 ($9$ labels), L2 ($11$ labels), and L3 ($22$ labels). As we progress down the levels, the features being annotated gradually progress from coarse to fine detail. The highest level (L1) contains classes that amalgamate several lower-level classes.
For example, \textit{Dense Regular Connective (C.D.R)} is an L3 precise label that falls under the more coarse L1 category of \textit{Connective (C)}. For the purpose of our work, we have selected L1 as it seems to be the most statistically significant selection with a better balance of per-class distribution.

\paragraph{ChestXray-14} 
The ChestX-Ray 14 dataset contains hospital-scale frontal-view chest X-ray images from 30,805 unique patients. Each image either contains multiple common thoracic illnesses including ‘cardiomegaly’ or ‘pneumonia’ or is designated ‘normal’ indicating no illness. The released version of the dataset catalogs 14 common illnesses to date, as opposed to the original 8 that was released at the time of publication.

\subsection{Hyperparameters \& Tuning}
In this section, we list all the necessary parameters for the reproducibility of our method. We have categorized our hyperparameters depending on which part of the pipeline they relate to (i.e., Training Optimization refers to any parameters used in setting up the training phase). A special note is made for the Loss Development $\lambda$ values. In order to best tune our method, we sampled a 15-point log-random search in a subset of the provided range to best adapt our model to the given datasets. See Table \ref{tab: hyperparameters}.

\begin{table*}
    \centering
    \caption{Details the hyperparameter settings used when training our models to ensure reproducibility.}
    \label{tab: hyperparameters}
    \resizebox{\textwidth}{!}{
    \normalsize{
    \begin{tabular}{c|c|c||c||c}
    \hline
    
    \multicolumn{3}{c||}{\textbf{Hyper-parameters}} &
    \multirow{2}{*}{\textbf{Range}} &
    \multirow{2}{*}{\textbf{Value}} \\
    \cline{1-3}
    
    \textbf{Category} & \textbf{Parameter Name} & \textbf{Description} & &\\
    \hline 
    \hline
    
    \multirow{4}{*}{\textbf{Training Optimization}}
    & \textbf{Base Learning rate} & Model Learning Rate & $(0, 1.0]$ & $0.0002$ \\ 
    \cline{2-5}

     & \textbf{Optimizer} &  Adam: Weight Decay = 1e-4 &-  &  -\\ 
    \cline{2-5}

     & \textbf{Schedueler} & One Cycle LR: PctStart = 0.2  & - & - \\ 
    \cline{2-5}

     & \textbf{ModelEMA} &  Exponential Moving average to encompass the model with decay: & $(0, 1.0]$ & $0.9997$ \\ 
    \cline{2-5}
    \hline
    \multirow{16}{*}{\textbf{Dataset Agumentations}}
    & \textbf{CutoutPIL} &   & $(0, 1.0]$ & $0.5$ \\ 
    \cline{2-5}
    & \multirow{14}{*}{\textbf{Default RandAugment \cite{cubuk2020randaugment}}} 
    & Shear - X & $(0, 0.3]$ & - \\ 
    && Shear - Y  & $(0, 0.3]$ &  -\\ 
     && Translate - X  & $(0, 150/331]$ & - \\ 
     && Translate - Y  & $(0, 150/331]$ &  -\\ 
     && Rotate   & $(0, 30]$ & - \\ 
     && Color   & $(0, 0.9]$ &  -\\ 
     && Posterize   & $(8, 4]$ &  -\\ 
     && Solarize   & $(256, 0]$ & -\\ 
     && Contrast   & $(0, 0.9]$ & - \\ 
     && Sharpness   & $(0, 0.9]$ & - \\ 
     && Brightness   & $(0, 0.9]$ & - \\
     && AutoContrast   & $[0]$ & $0$ \\
     && Equalize  & $[0]$ & $0$ \\
     && Invert  & $[0]$ & $0$ \\
    
    \cline{2-5}
    & \textbf{Normalization} &  Applied Mean and Standard Deviation based on Computer Vision and Medical Dataset Respectively & -& - \\ 
    \cline{2-5}
    \hline
    \multirow{8}{*}{\textbf{Loss Development}}
     & \textbf{$\lambda_1$} & Controls the contribution of the KMCL loss  & $(0, 1.0]$ & $0.3$ \\ 
    \cline{2-5}

     & \textbf{$\lambda_2$} &  Controls the contribution of the ASL loss  & $(0, 1.0]$ & $0.1$ \\ 
    \cline{2-5}

     & \textbf{$\gamma_{+}$} & Internal Parameter of the ASL Loss  & $-$ & $0$ \\ 
    \cline{2-5}

    & \textbf{$\gamma_{-}$} &  Internal Parameter of the ASL Loss & $-$ & $4$ \\ 
    \cline{2-5}

    & \textbf{$m$} & Internal Parameter of the ASL Loss  & $(0, -]$ & $0.05$ \\ 
    \cline{2-5}

     & \textbf{$\tau$} & Temperature Scaling factor for the KMCL loss function  & $(0, 1.0]$ & $0.2$ \\ 
    \cline{2-5}

    & \textbf{$\epsilon$} & ELU epsilon factor for reconstruction loss stability  & $(0, 1.0]$ & $1e-7$ \\ 
    \cline{2-5}

    & \textbf{$\alpha$} & ELU alpha factor for reconstruction loss stability  & $(0, 1.0]$ & $1$ \\ 
    \cline{2-5}

     \hline
    \multirow{6}{*}{\textbf{KMM Layer Initialization}}
     & \textbf{Mixture Coefficient Weights*} & Initialization was sampled from uniform distribution  & $(0, 0.1]$ & - \\ 
    \cline{2-5}

     & \textbf{Mean Weights} & Initialization was sampled from uniform distribution  & $(0, 0.1]$ & - \\ 
    \cline{2-5}

     & \textbf{Variance Weights} & Initialization was constant 1  & $(0, 1.0]$ & $1$ \\ 
    \cline{2-5}
    & \textbf{Mixture Coefficient Bias} & Initialization was constant 0  & $(0, 1.0]$ & $0$\\ 
    \cline{2-5}

     & \textbf{Mean Bias} & Initialization was constant 0  & $(0, 1.0]$ & $0$ \\
    \cline{2-5}

     & \textbf{Variance Bias} & Initialization was constant 0  & $(0, 1.0]$ & $0$ \\ 
    \cline{2-5}
    
    \hline
    
    \end{tabular}
    }
    }
\end{table*}

\subsection{Additional Information on Metrics}
Being consistent with state-of-the-art methods, we calculate the average overall precision (OP), recall (OR), and F1 score (OF1), in addition to the average per-class precision (CP), recall (CR), and F1 score (CF1), as metrics for evaluating the different methods on the datasets \cite{chen2019multi,ridnik2021asymmetric,zhao2021transformer}. Overall these metrics challenge the model’s ability to accurately discriminate the class of interest in terms of measuring false positives and false negatives. Superior OF1 and CF1 indicate that the model is well-tuned for class discrimination as this metric encompasses both recall and precision in the calculation. For some experiments, we include the following computational complexity measures: Parameters (MM) to indicate model size, and GMAC to indicate the forward computational resource required. The motivation behind these metrics is to illustrate that performance is not only measured through how well the method discriminates classes but also through the complexity of deploying said method in the real world. Finally, due to the increased difficulty of the ChestX-ray14 dataset, we additionally report per class AUC scores to identify model discriminability for the class of interest, this has been a common trend in papers that have cited results on this dataset \cite{cohen2020limits, rajpurkar2017chexnet}.

\subsection{Additional Visualizations}
\begin{figure}
    \centering
    \includegraphics[width=1.0\textwidth]{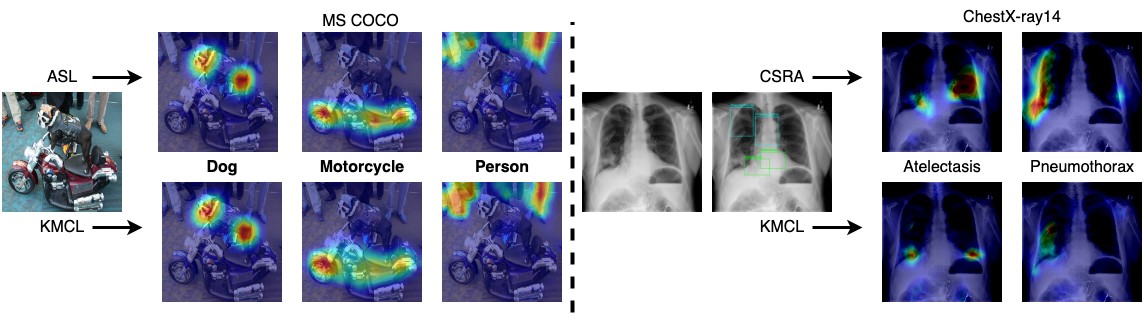}
    \caption{GradCam Visualization of KMCL against top competitive method on the MS-COCO and ChestXray-14 Dataset.}
    \label{fig:addGradCam}
\end{figure}

To further augment the main paper visualizations, we attach supplemental visualizations on the two additional datasets: MS-COCO and ChestXray-14. As can be seen, by the visualizations, our model is more precise at localizing the correct features. Due to capturing the epistemic uncertainty from the kernel representation, our method is able to focus the activation on the correct class, limiting extraneous false positive results. See Figure \ref{fig:addGradCam}.

\end{document}